\documentclass[10pt,twocolumn,letterpaper]{article}

\usepackage[pagenumbers]{cvpr} 

\usepackage{graphicx}
\usepackage{amsmath}
\usepackage{amssymb}
\usepackage{booktabs}
\usepackage[accsupp]{axessibility}

\usepackage[pagebackref,breaklinks,colorlinks]{hyperref}

\usepackage[capitalize]{cleveref}
\crefname{section}{Sec.}{Secs.}
\Crefname{section}{Section}{Sections}
\Crefname{table}{Table}{Tables}
\crefname{table}{Tab.}{Tabs.}

\def\cvprPaperID{6655} 
\def\confName{CVPR}
\def\confYear{2023}

\begin{document}

\title{GeoVLN: Learning Geometry-Enhanced Visual Representation with Slot Attention for Vision-and-Language Navigation}

\author{Jingyang Huo\footnotemark[1], Qiang Sun\footnotemark[1], Boyan Jiang\footnotemark[1], Haitao Lin, Yanwei Fu\footnotemark[2]\\
Fudan University\\
}

\maketitle

\renewcommand{\thefootnote}{\fnsymbol{footnote}} 
\footnotetext[1]{Equal contributions.}
\footnotetext[2]{Corresponding authors.}
\footnotetext{Yanwei Fu is with School of Data Science, Fudan University, Shanghai Key Lab of Intelligent Information Processing, and Fudan ISTBI–ZJNU Algorithm Centre for Brain-inspired Intelligence, Zhejiang Normal University, Jinhua, China.}

\begin{abstract}
Most existing works solving Room-to-Room VLN problem only utilize RGB images and do not consider local context around candidate views, which lack sufficient visual cues about surrounding environment. Moreover, natural language contains complex semantic information thus its correlations with visual inputs are hard to model merely with cross attention. In this paper, we propose GeoVLN, which learns \textbf{Geo}metry-enhanced visual representation based on slot attention for robust \textbf{V}isual-and-\textbf{L}anguage \textbf{N}avigation. The RGB images are compensated with the corresponding depth maps and normal maps predicted by Omnidata as visual inputs. Technically, we introduce a two-stage module that combine local slot attention and CLIP model to produce geometry-enhanced representation from such input. We employ V\&L BERT to learn a cross-modal representation that incorporate both language and vision informations. Additionally, a novel multiway attention module is designed, encouraging different phrases of input instruction to exploit the most related features from visual input. Extensive experiments demonstrate the effectiveness of our newly designed modules and show the compelling performance of the proposed method. Code and models are available at \url{https://github.com/jingyanghuo/GeoVLN}.
\end{abstract}

\section{Introduction}
\label{sec:intro}

\begin{figure}
\begin{centering}
\includegraphics[width=0.99\linewidth]{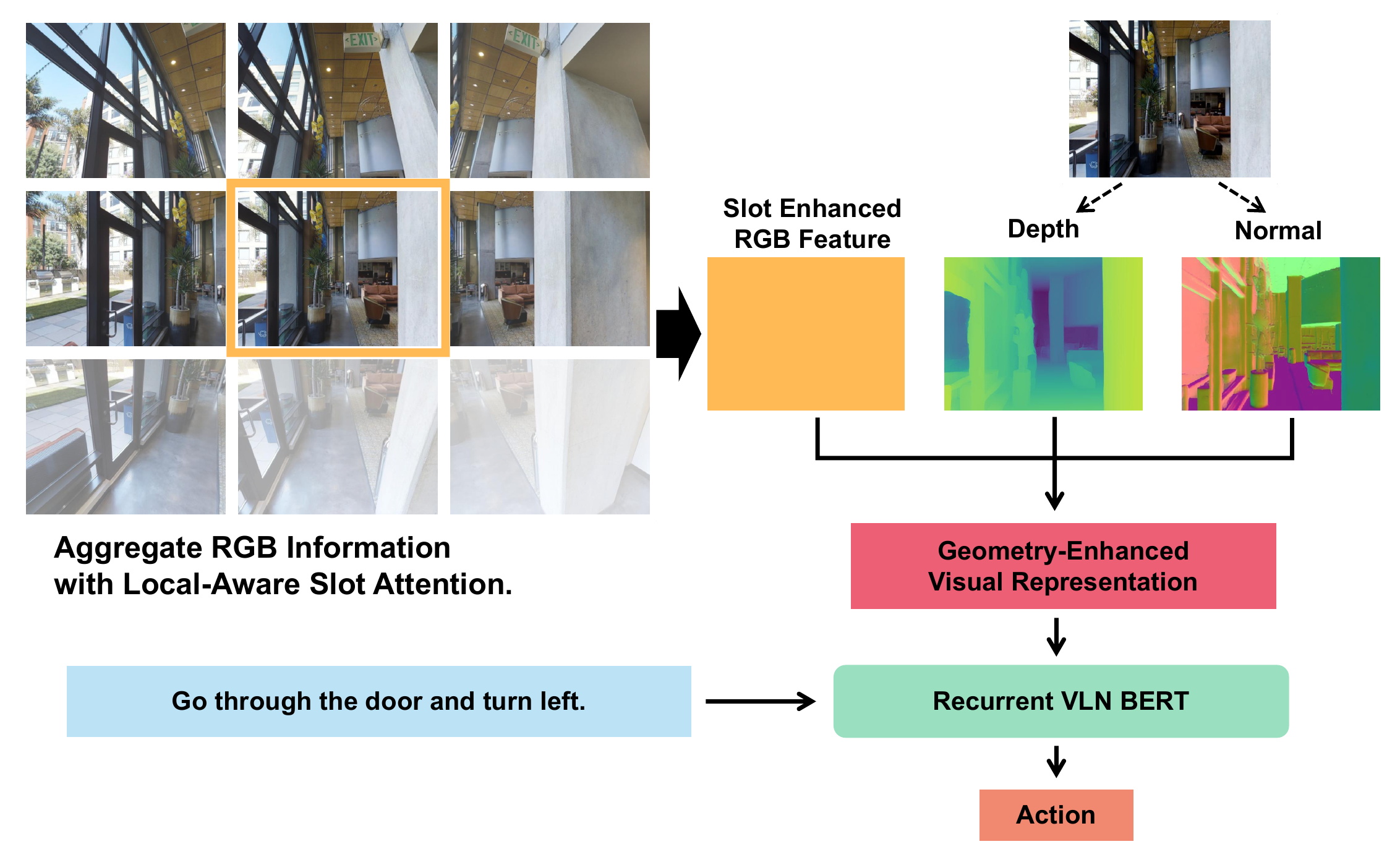}
\par\end{centering}
 \vspace{-0.1in}
 \caption{Illustration of our learning geometry-enhanced visual representation (GeoVLN) for visual-and-language navigation. Critically, our GeoVLN utilizes the slot attention mechanism.
 \label{fig:teaser}}
 \vspace{-0.15in}
\end{figure}

With the rapid development of vision, robotics, and AI research in the past decade, asking robots to follow human instructions to complete various tasks is no longer an unattainable dream. To achieve this, one of the fundamental problems is that given a natural language instruction, let robot (agent) make its decision about the next move automatically based on past and current visual observations. This is referred as Vision-and-Language Navigation (VLN)~\cite{R2R}. Importantly, such navigation abilities should also work well in previously unseen environments.

In the popular Room-to-Room navigation task \cite{R2R}, the agent is typically assumed to be equipped with a single RGB camera. At each time step, given a set of visual observations captured from different view directions and several navigation options, the goal is to choose an option as the next station. The process will be repeated until the agent reaches the end point described by the user instruction. Involving both natural language and vision information, the main challenge here is to learn a cross-modal representation that incorporates the correlations between user instruction and current surrounding environment to aid decision-making.

As solutions, early studies~\cite{R2R, HaoTan2019LearningTN, DanielFried2018SpeakerFollowerMF} resort to LSTM \cite{hochreiter1997long} to process temporal visual data stream. 
However, recent works \cite{YuankaiQi2021TheRT, XiujunLi2019RobustNW, ArjunMajumdar2022ImprovingVN, hao2020towards, hong2021vln, HAMT, XiangruLin2021SceneIntuitiveAF} have taken advantage of the superior performance of the Transformer \cite{AshishVaswani2017AttentionIA} and typically employ this attention-based model to facilitate representation learning with cross attention and predict actions in either recurrent \cite{hong2021vln} or one-shot \cite{HAMT} fashion. Despite their advantages, these approaches still have several limitations.

\begin{itemize}
  
 \item  1) They only rely on RGB images which provide very limited 2D visual cues and lack geometry information. Thus it is hard for agent to build scene understanding about novel environments; 
\item 2) they process each candidate view independently without considering local spatial context, leading to inaccurate decisions; 
\item  3) natural language contains high-level semantic features and different phrases within an instruction may focus on various aspects visual information, \eg texture, geometry. Nevertheless, we empirically find that constructing cross-modal representation with na\"ive attention mechanism leads to suboptimal performance.

\end{itemize}

To address these problems, we propose a novel framework, named GeoVLN, which learns \textbf{Geo}metry-enhanced visual representation based on slot attention for robust \textbf{V}isual-and-\textbf{L}anguage \textbf{N}avigation. Our framework is illustrated in Fig.~\ref{fig:teaser}. In particular, beyond RGB images, we also utilize the corresponding depth maps and normal maps as observations at each time step (Fig.~\ref{fig:teaser}), as they provide rich geometry information about environment that facilitates decision-making. Crucially, these additional mid-level cues are estimated by the recent scalable data generation framework Omnidata \cite{eftekhar2021omnidata, kar20223d} rather than sensor captured or user provided.

We design a novel two-stage slot attention \cite{slotatt} based module to learn geometry-enhanced visual representation from the above multimodal observations. Note that the slot attention is originally proposed to learn object-centric representation for complex scenes from single/multi-view images, 
but we utilize its feature learning capability and extend it to work together with multimodal observations in the VLN tasks. Particularly, we treat each candidate RGB image as a query, and choose its nearby views as keys and values to perform slot attention within a local spatial context. The key insight is that our model can implicitly learn view-to-view correspondences via slot attention, and thus encourage the candidates to pool useful features from surrounding neighbors. 
Additionally, we process all complementary observations, including depth maps and normal maps, through a pre-trained CLIP~\cite{AlecRadford2021LearningTV} image encoder to obtain respective latent vectors.
These vectors are then concatenated with the output of slot attention module to form our final geometry-enhanced visual representation.

On the other hand, we employ BERT as language encoder to acquire global latent state and word embeddings from the input instruction. Given the respective latent embeddings for language and vision inputs, we adopt V\&L BERT \cite{hao2020towards} to merge multimodal features and learn cross-modal representation for the final decision-making in a recurrent fashion following \cite{hong2021vln}. Different from previous works \cite{ChihYaoMa2019SelfMonitoringNA, HaoTan2019LearningTN} that directly output probabilities of each candidate option, we present a multi-way attention module to encourage different phrases of input instruction to focus on the most informative visual observation, which boosts the performance of our network, especially in unseen environments.

To summarize, we propose the following contributions that solve the Room-to-Room VLN task with compelling accuracy and robustness:

\begin{itemize}
\item 
We extend slot attention to work on VLN task, which is combined with CLIP image encoder to learn geometry-enhanced visual representations for accurate and robust navigation.

\item 
A novel multiway attention module encouraging different phrases of input instruction to focus on the most informative visual observation, \eg texture, depth.

\item 
We compensate RGB images with the corresponding depth maps and normal maps predicted with off-the-shelf method, improving the performance yet not involving additional training data.

\item 
We integrate all the above technical innovations into a unified framework, named GeoVLN, and the extensive experiments validate our design choices.
\end{itemize}
\section{Related Work}
\label{sec:related}

\noindent
{\bf Vision-and-Language Navigation}
Exploring agents that can follow instructions to navigate in real-world scenarios is a challenging yet crucial research area for embodied artificial intelligence~\cite{9687596}.
Promoted by the recent proposed large-scale datasets, including Matterport3D~\cite{Matterport3D}, Habitat~\cite{FeiXia2018GibsonER}, Gibson~\cite{ManolisSavva2019HabitatAP} and Room-to-Room~\cite{R2R}, agent navigation tasks can be performed in photorealistic environments.
Researchers have proposed solutions in terms of intra-modal modeling, cross-modal alignment and decision-making learning, respectively~\cite{YicongHong2020LanguageAV, ChihYaoMa2019SelfMonitoringNA, XinWang2018ReinforcedCM}. Early studies~\cite{R2R, HaoTan2019LearningTN, DanielFried2018SpeakerFollowerMF} use LSTM as the backbone. Benefiting from the superior performance of BERT~\cite{devlin2018bert}, a series of recent approaches based on pre-trained vision-and-language (V\&L) BERT~\cite{YuankaiQi2021TheRT, XiujunLi2019RobustNW, ArjunMajumdar2022ImprovingVN, hao2020towards, hong2021vln, HAMT, XiangruLin2021SceneIntuitiveAF} are introduced to the VLN task and outperform the traditional LSTM-based baselines. To name a few, PREVALENT ~\cite{hao2020towards} conducts V\&L BERT pre-training for the image-text-action triplets and firstly derives generic representations of visual and linguistic clues applicable to VLN. Hong~\etal~\cite{hong2021vln} augments V\&L BERT with a recurrent function to model the time-dependent information in the navigation process. HAMT~\cite{HAMT} introduces a hierarchical vision transformer to capture the spatial and temporal relationships of historical observations separately, thus benefits the decision-making process. 
Different from prior arts \cite{hong2021vln,HAMT} that directly predict next action based on current multimodal conditions, we present a novel multiway attention module inspired by MAttNet \cite{LichengYu2018MAttNetMA} to learn the correlations between the user instruction and different modalities of input observation, improving the accuracy of final decision.

\noindent
{\bf Visual Representation in VLN}
In VLN task, directly training the whole framework end-to-end on high-resolution images is usually infeasible due to the expensive memory and computation cost. To tackle this problem, prior works \cite{R2R,SimoneParisi2022TheE} typically utilize perceptual feature precomputed by the large pretained extractors, \eg ResNet \cite{he2016deep}, and demonstrate its effectiveness.
Recently, MURAL-large~\cite{AashiJain2021MURALMM} and CLIP~\cite{AlecRadford2021LearningTV, ShengShen2021HowMC} show their capability of learning expressive representations across multimodal data which facilitate several downstream tasks. We use CLIP to extract visual representation, since its feature space captures information shared by both vision and text that may encourage our model to learn semantic correspondences between them.

Additionally, some recent works \cite{yu2021unsupervised,sajjadi2022object} employ slot attention \cite{slotatt} to learn high-level object-centric representation from single/multi-view images of a complex scene, and achieve impressive neural rendering results. Zhuang~\etal~\cite{local} adopt vanilla slot attention module to aggregate object-centric visual information spatially. 
Different from~\cite{local}, we newly design a novel slot-based visual representation learning module that aggregates the information from both spatial neighbors and multiple visual modalities, so that construct a better understanding of the surrounding environment for agent.

\section{Method}
\label{method}

\begin{figure*}
\begin{centering}
\vspace{-0.23in}
\includegraphics[width=0.86\linewidth]{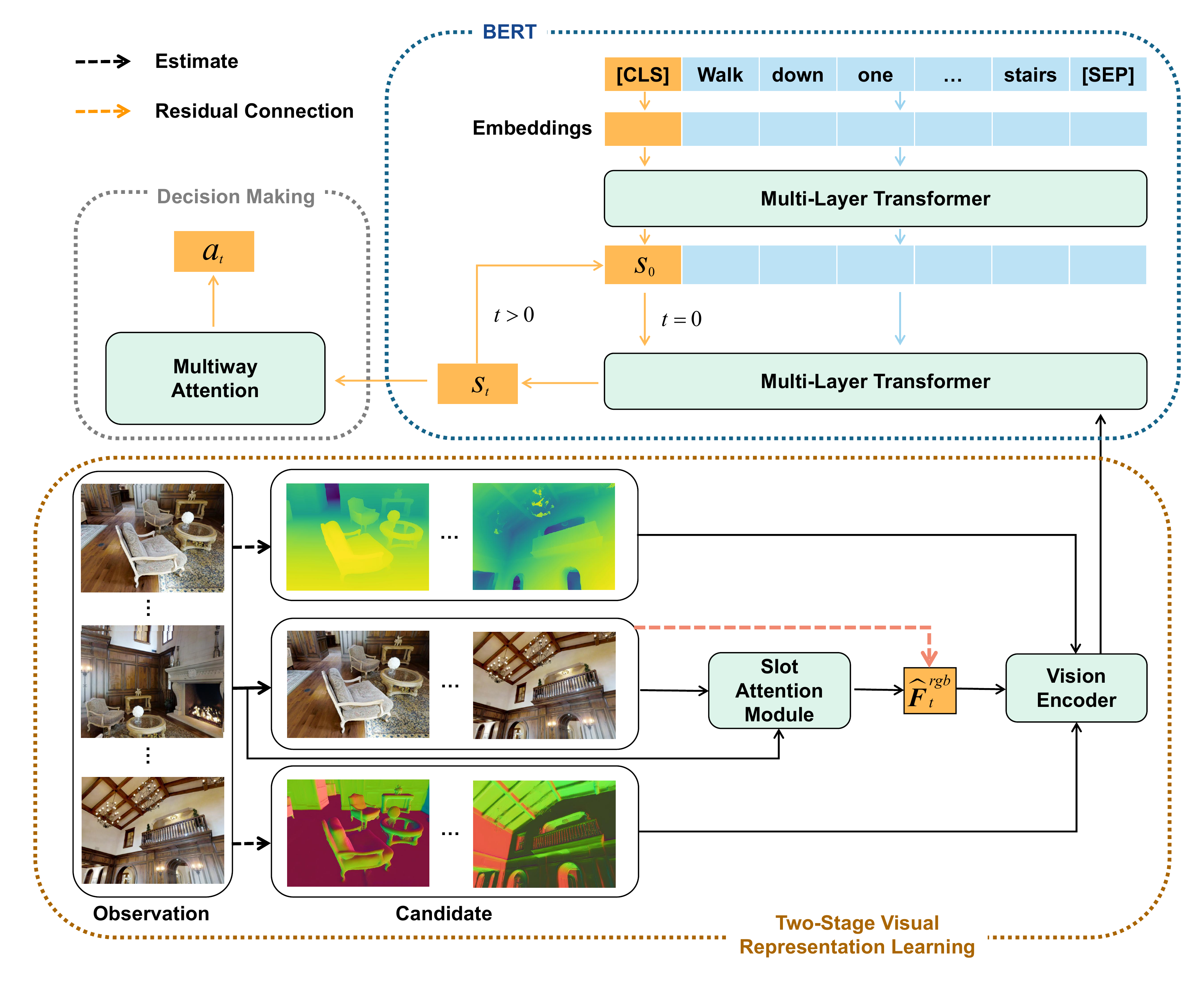}
\par\end{centering}
\vspace{-0.2in}
 \caption{The overview of our GeoVLN. Particularly, at each time step, our GeoVLN takes a single user instruction and a set of visual observations as input. The language input is consumed by BERT encoder  to obtain a global latent state  and a sequence of work embeddings. The visual input is composed of 36-view RGB images and the corresponding depth maps and normal maps estimated with Omnidata. We have a two-stage module to process such multimodal observations and acquire a geometry-enhanced visual representation.\label{fig:overview}}
\vspace{-0.15in}
\end{figure*}

\noindent
{\bf Overview}\quad
The pipeline of GeoVLN is overviewed in \cref{fig:overview}. 
At each time step $t$, our framework takes a single user instruction and a set of visual observations as input. The language input is consumed by BERT encoder \cite{devlin2018bert} to obtain a global latent state $\boldsymbol{s}_0$ and a sequence of work embeddings. The visual input is composed of 36-view RGB images and the corresponding depth maps and normal maps estimated with Omnidata \cite{eftekhar2021omnidata}. We design a two-stage module (\cref{sec:visual_repr}) to process such multimodal observations and acquire a geometry-enhanced visual representation. Given both language and vision representations, the final action is predicted by a multiway attention based decision making module introduced in \cref{sec:multiway_att}. We detail the loss functions used during training in \cref{sec:loss_func}.

\subsection{Problem Setups}
The Visual Language Navigation task in a discrete environment is defined on a preset connectivity graph $\mathcal{G}=\{\mathcal{V}, \mathcal{E}\}$, where $\mathcal{V}$ and $\mathcal{E}$ denote the vertex set and edge set of the connected graph, respectively. The agent, placed at an arbitrary starting position, is asked to follow user instructions to move between vertices along the edges of the connected graph until reaching the target destination. This navigation process can be formulated concretely as follows.

Firstly, the agent is placed at a starting point in a navigable environment described by the connectivity graph. There are many different viewpoints (the number depends on specific scene) the agent can station.
Then, given an instruction $\boldsymbol{I}$ containing a sequence of words,
the agent is required to approach the target following the instruction. At each time step $t$, the agent receives observations $\boldsymbol{O}_t$ as the visual input and makes its decision about the action $a_t$ to shift from the current state $s_t$ to the next state $s_{t+1}$.
It is worth noting that the observation $\boldsymbol{O}_t=\{o_t^{(i)}\}_{i=1}^{36}$ are $36$ perspective projection images from different view directions rather than a complete panoramic image. 
These view directions have horizontal angles sampled with $30^{\circ}$ intervals from $0^{\circ}-360^{\circ}$, and pitch angles chosen from $\left[-30^{\circ},0^{\circ},30^{\circ}\right]$.
At each viewpoint, the agent is also provided with $K$ candidate views $\boldsymbol{C}_t=\{c_t^{(i)}\}_{i=1}^{K}\subset \boldsymbol{O}_t$, corresponding to the navigable directions on the connectivity graph.
The output action that the agent decided at each time step is restricted to either one of the candidate views $\boldsymbol{C}_t$ or a special ``STOP'' signal, denote moving to the corresponding viewpoint or decide to stop.

\subsection{Two-Stage Visual Representation Learning} \label{sec:visual_repr}

\noindent \textbf{Visual Observations}
Most of the previous works only exploit RGB images as the visual observations in VLN task. However, such limited cues may involve biases about color information and lead to overfitting problem on the training environments so that hinder the generalization capability to novel scenes.
To alleviate this problem, we involve other data modalities of depth maps and surface normal maps as compensation to provide geometry information that is nontrivial to be directly obtained from RGB images.
Crucially, the depth maps and normal maps are estimated with the recent proposed Omnidata ~\cite{eftekhar2021omnidata} without any additional training data. 

We employ CLIP image encoder \cite{AlecRadford2021LearningTV} pretrained on the large-scale dataset of image-text pairs to extract feature vectors (640-dimension) from all the visual observations, which are denoted as $\boldsymbol{O}_t^{rgb}$, $\boldsymbol{O}_t^{dep}$ and $\boldsymbol{O}_t^{nor}$ for RGB image, depth map and normal map respectively. We use $\{\boldsymbol{C}_t^{\ast} \mid {\ast} \in \left[rgb, dep, nor\right]\}$ to refer to the corresponding features of candidate views.
Additionally, given the view angles $\{\theta,\varphi\}$ of each candidate image, we obtain an angle embedding $\boldsymbol{F}_t^{ang}$ by repeating $\left(\sin \left(\theta_i\right), \cos \left(\theta_i\right), \sin \left(\varphi_i\right), \cos \left(\varphi_i\right)\right)$ 32 times as typically used in previous work \cite{hong2021vln}. We concatenate the visual features and angle features together to obtain the final candidate features:
\begin{equation}
    \boldsymbol{F}_t^{\ast} = [ \boldsymbol{C}_t^{\ast}; \boldsymbol{F}_t^{ang} ], \quad {\ast} \in \left[rgb, dep, nor\right].
\end{equation}
And the features of the 36 views RGB observations (\ie panoramic views) are: 
\begin{equation}
    \boldsymbol{P}_t^{rgb} = [ \boldsymbol{O}_t^{rgb}; \boldsymbol{F}_t^{ang} ],
\end{equation}
which are fed into the slot attention module introduced below to fuse information from local neighboring views.

\noindent \textbf{Local-Aware Slot Attention}
Some recent works \cite{hong2021vln, hao2020towards, HaoTan2019LearningTN} only utilize the candidate views during navigation process. This brings the obstacle of understanding surround environment so that hinder navigation accuracy. For example, there are very few candidates at some viewpoints that are insufficient for making decision about next move.
To mitigate this problem, we employ a slot attention module to encourage each candidate views $\boldsymbol{C}_t$ to aggregate information from the nearby observation views $\boldsymbol{O}_t$ according to the spatial proximity principle.
Specifically, we initialize the slots with RGB candidate features $\boldsymbol{F}_t^{rgb}$ and treat them as queries when performing attention calculation.
Additionally, the observation features $\boldsymbol{P}_t^{rgb}$ and $\boldsymbol{O}_t^{rgb}$ are used as keys and values.
We apply a dropout layer to the inputs:
\begin{equation}
\begin{split}
\text{slots} &=\operatorname{Dropout}(\boldsymbol{F}_t^{rgb}),\\
Q &=\operatorname{Dropout}(\operatorname{LN}(\text{slots})),\\
K &=\operatorname{Dropout}(\operatorname{LN}(\boldsymbol{P}_t^{rgb})),\\
V &=\operatorname{Dropout}(\operatorname{LN}(\boldsymbol{O}_t^{rgb})),
\end{split}
\end{equation}
where LN denotes layer normalization.

\begin{figure}
\begin{centering}
\includegraphics[width=0.6\linewidth]{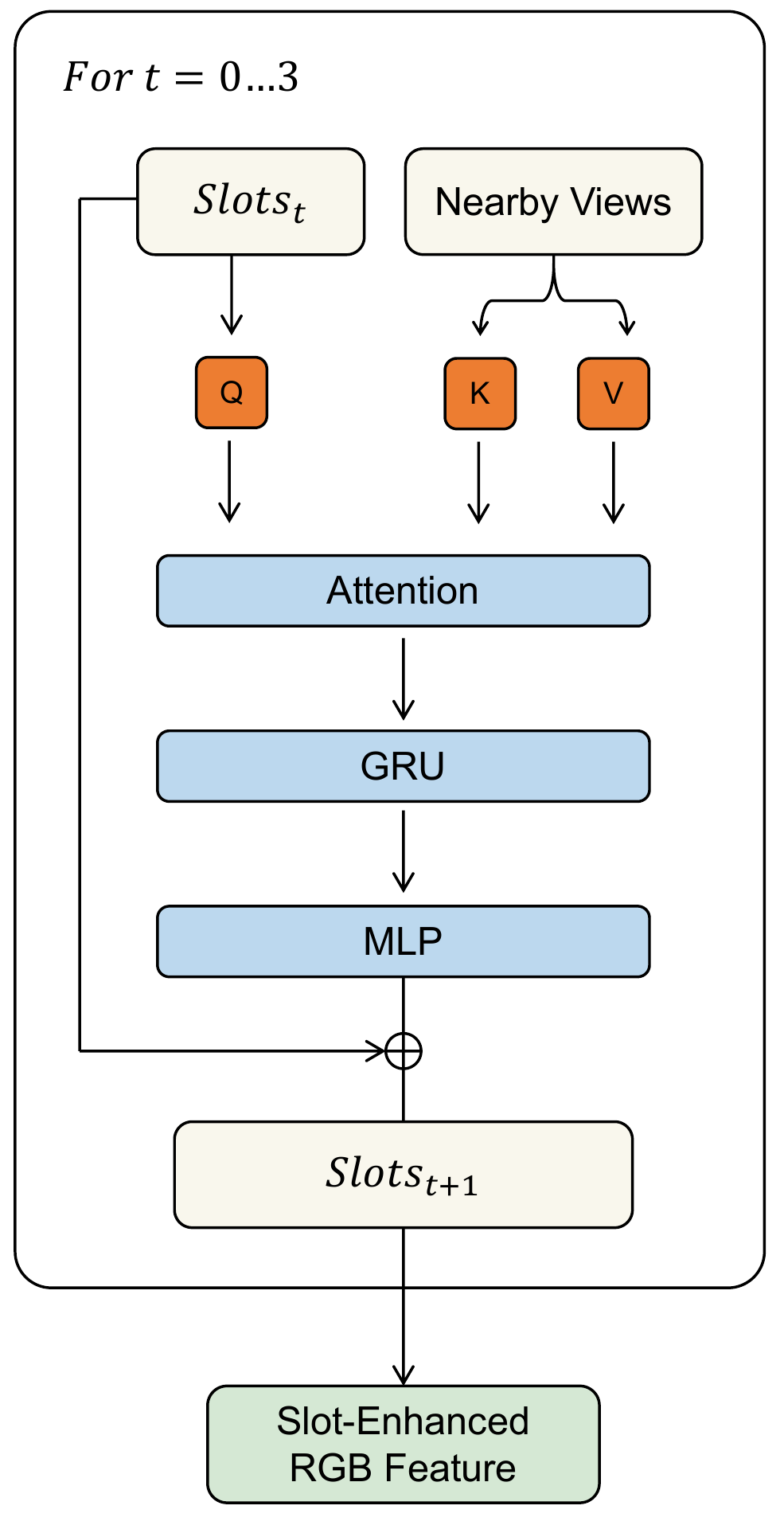}
\par\end{centering}
\vspace{-0.1in}
 \caption{Detailed architecture of our local-aware slot attention module. \label{fig:slotatt}}
\vspace{-0.1in}
\end{figure}

As shown in \cref{fig:slotatt}, the slots are updated in a recurrent fashion following~\cite{slotatt}. At each updating step $t=1, \cdots, T$ ($T=3$ in our experiments), we compute dot-product attention between keys and queries as the widely-used cross-attention, while apply Softmax operator along slot dimension to normalize the attention scores, which forces the candidate views to competitively access the information of the observations $\boldsymbol{O}_t$: 
\begin{align}
\text{updates} &=\operatorname{Softmax}\left(\frac{Q \cdot K^{\top}}{\sqrt{d_Q}}, \text{axis=slot} \right)V,
\end{align}
where $d_Q$ is the dimension of $Q$. 
Then the slots are updated with a Gated Recurrent Unit (GRU) followed by a residual MLP: 
\begin{equation}
\begin{split}
\text{slots} &= \operatorname{GRU} (\text{state=slots}, \text{inputs=updates}),\\
\text{slots} &= \text{slots} + \operatorname{MLP}(\operatorname{LN}((\text{updates})).
\end{split}
\end{equation}
With slot attention, the representation of each candidate view is progressively refined based on the observations $\boldsymbol{O}_t^{rgb}$, so that the agent can capture more information from a single viewpoint to aid decision making. However, directly using all observations at one viewpoint would involve non-local information and hinder the convergence of our model. Therefore, we restrict each candidate view to focus on observations whose heading and elevation angles differ by no more than $30^{\circ}$ from itself. We achieve this by using attention masks.

As shown in \cref{fig:overview}, we use a residual connection to add the updated slots to the candidate features $\boldsymbol{F}_t^{rgb}$. Note that we only update the visual features and keep the angle features fixed: 
\begin{equation}
    \label{eq:slot}
    \hat{\boldsymbol{F}}_t^{rgb} = \left[ \boldsymbol{C}_t^{rgb} + \text{slots}[..., :d_C]; \boldsymbol{F}_t^{ang} \right],
\end{equation}
where $d_C$ is the dimension of $\boldsymbol{C}_t^{rgb}$.

The output of our local-aware slot attention module is denoted as $\hat{\boldsymbol{F}}_t^{rgb}$.
We then concatenate $\hat{\boldsymbol{F}}_t^{rgb}$ with the visual features from depth map and normal map, together with view angle feature, and project it into a 768-dimensional vector with a fully connected layer followed by a layer normalization. 
\begin{equation}
\begin{split}
\boldsymbol{F}_t &= \left[ \hat{\boldsymbol{F}}_t^{rgb}[..., :d_C]; 
 \boldsymbol{C}_t^{dep}; \boldsymbol{C}_t^{nor}; \boldsymbol{F}_t^{ang} \right] \\
\hat{\boldsymbol{F}}_t &=\operatorname{LN}(\operatorname{FC}(\boldsymbol{F}_t))
\end{split}
\end{equation}

The resulting geometry-enhanced visual representation $\hat{\boldsymbol{F}}_t$ will be used as the visual tokens of the Recurrent VLN BERT.

\subsection{Multiway Attention Based Decision Making} \label{sec:multiway_att}
\noindent \textbf{Recurrent VLN BERT}
We adopt Recurrent VLN-BERT to process the instruction $\boldsymbol{I}$ and the geometry-enhanced visual representation $\hat{\boldsymbol{F}}_t$ obtained from slot attention module. At each time step, the global state vector $\boldsymbol{s}_t$ and tokens are updated with a multi-layer Transformer, which can be formulated as:
\begin{equation}
\boldsymbol{s}_t=\operatorname{VLN} \circlearrowright \operatorname{BERT}\left(\boldsymbol{s}_{t-1}, \boldsymbol{I}, \hat{\boldsymbol{F}}_t\right).
\end{equation}
Note that $\boldsymbol{s}_t$ contains information of both vision, language as well as all the past decisions of the agent, we utilize it as the cross-modal representation to support subsequent decision-making.

\vspace{0.8em}
\noindent
{\bf Multiway Attention}
Different from previous works which output decisions directly from multi layer Transformer, we design a multiway attention module to compute attention scores of $\boldsymbol{s}_t$ with three modalities of visual observations: RGB, depth and normal individually, and obtain the final policy likelihood by weighted summation.
We take the attention calculation with RGB features as an example. Firstly, the state representation $\boldsymbol{s}_t$ is directly projected into a 768-dimensional latent vector, while the RGB features $\hat{\boldsymbol{F}}_t^{rgb}$ are normalized by a LayerNorm (LN) operation and then projected to the same dimension through a fully connected (FC) layer:
\begin{equation}
\begin{split}
\tilde{\boldsymbol{s}}_t^{rgb}&=\boldsymbol{s}_t \boldsymbol{W}^{s,rgb}, \\
\tilde{\boldsymbol{F}}_t^{rgb}&=\operatorname{FC}(\operatorname{LN}(\hat{\boldsymbol{F}}_t^{rgb})).
\end{split}
\end{equation}
Then, the attention score can be computed as: 
\begin{equation}
\boldsymbol{A}_t^{rgb}=\frac{\tilde{\boldsymbol{F}}_t^{rgb} \tilde{\boldsymbol{s}}_t^{rgb}{ }^{\top}}{\sqrt{d_h}},
\end{equation}
where $d_h$ denotes the dimension of the hidden space. Similarly, the attention scores $\boldsymbol{A}_t^{dep}$ and $\boldsymbol{A}_t^{nor}$ can be obtained for the depth features and the normal features, respectively.

\vspace{0.8em}
\noindent
{\bf Matching Score}\quad
At each time step, how much each modality contributes to the navigation should differ noticeably. For instance, the agent may focus more on the depth information when executing the instruction ``go through the corridor'' whereas the process of ``picking up the spoon'' will be more pertinent to the RGB and normal information. To achieve this, we apply a fully connected layer followed by a Softmax operation to compute the weights corresponding to the three modalities: 
\begin{equation}
\left[w_t^{rgb}, w_t^{dep}, w_t^{nor}\right]=\operatorname{Softmax}\left( \boldsymbol{s}_t \boldsymbol{W}^m+\boldsymbol{b}^m\right),
\end{equation}
where $\boldsymbol{W}^m$ and $\boldsymbol{b}^m$ are learnable parameters.

Thus, the final matching scores of the candidate views $\boldsymbol{C}_t$ w.r.t. the state vector $\boldsymbol{s}_t$ can be written as:
\begin{equation}
\begin{split}
 \boldsymbol{S}_t^{total} &= w_t^{rgb} \boldsymbol{A}_t^{rgb} + w_t^{dep} \boldsymbol{A}_t^{dep} + w_t^{nor} \boldsymbol{A}_t^{nor}, \\
\end{split}
\end{equation}
where $\boldsymbol{S}_t^{total}$ is a $(K+1)$-dimensional vector and $K$ denotes the number of candidate views at current viewpoint. We use $\boldsymbol{S}_{t, i}^{total} (1 \leq i \leq K)$ to denote the matching score of the $i$-th candidate view, while $\boldsymbol{S}_{t, K+1}^{total}$ denotes the matching score of the ``STOP'' action. 
We denote the action probabilities as $\boldsymbol{p}_t$ obtained by applying the softmax function to $\boldsymbol{S}_t^{total}$. 
The candidate view with the highest probability is then selected as the final decision.

\subsection{Loss Function} \label{sec:loss_func}

We follow the training protocol used in~\cite{hong2021vln}, which combines imitation learning and reinforcement learning. Specifically, our objective functions is composed of two parts. The first part is the cross-entropy loss derived from the teacher-forcing method \cite{williams1989learning}. The teacher actions are determined by the human-labeled ground-truth trajectories. Denoting the teacher action as $a^{\ast}$, the loss of imitation learning can be formulated as: 
\begin{equation}
\mathcal{L}_{IL}= - \sum_t a_t^{\ast} \log \left(\boldsymbol{p}_t\right).
\end{equation}

Secondly, we use the A2C~\cite{VolodymyrMnih2016AsynchronousMF} algorithm identical to the one set in ~\cite{hong2021vln}. 
At each time step, an action is sampled according to $\boldsymbol{S}_t^{total}$ and a reward strategy is applied following the set-up. 
The reinforcement learning loss (\cref{rl}) is composed of three components: an actor loss to optimize strategy, a critic loss to estimate the state vector, and a regular loss to reduce action uncertainty. Additional details can be found in~\cite{VolodymyrMnih2016AsynchronousMF,hong2021vln}.
\begin{equation}
\label{rl}
\mathcal{L}_{RL}=\sum_t \mathcal{L}_{\text {actor }}^{(t)}+\mathcal{L}_{\text {critic }}^{(t)}+\lambda_{\text {reg }} \mathcal{L}_{\text {reg }}^{(t)}
\end{equation} 

The overall objective function guiding our training process is
\begin{equation}
\mathcal{L}=\mathcal{L}_{RL} + \lambda \mathcal{L}_{IL},
\end{equation}
where $\lambda$ denotes the loss weight to balance both terms.


\begin{table*} \small 
    \centering
    \renewcommand\arraystretch{1.0}
    \scalebox{1.0}{
    \begin{tabular}{lcccccccccccc}
        \hline
         \toprule
         \multicolumn{1}{c}{} & \multicolumn{4}{c}{Val Seen} & \multicolumn{4}{c}{Val Unseen} & \multicolumn{4}{c}{Test Unseen} \\
         \cmidrule(lr){2-5} \cmidrule(lr){6-9} \cmidrule(lr){10-13} 
         Agent & TL$\downarrow$ & NE$\downarrow$ & SR$\uparrow$ & SPL$\uparrow$ & TL$\downarrow$ & NE$\downarrow$ & SR$\uparrow$ & SPL$\uparrow$ & TL$\downarrow$ & NE$\downarrow$ & SR$\uparrow$ & SPL$\uparrow$ \\
         \hline
         RANDOM~\cite{R2R} & 9.58 & 9.45 & 16 & - & 9.77 & 9.23 & 16 & - &  9.93 & 9.77 & 13 & 12 \\
         Human& - & - & - & - & - & - & - & - & 11.85 & 1.61 & 86 & 76 \\
         \hline
         Seq-to-Seq~\cite{R2R}& 11.33 & 6.01 & 39 & - & \textbf{8.39} & 7.81 & 22 & - & \textbf{8.13} & 7.85 & 20 & 18 \\
         Speaker-Follower~\cite{DanielFried2018SpeakerFollowerMF} & - & 3.36 & 66 & - & - & 6.62 & 35 & - & 14.82 & 6.62 & 35 & 28 \\
         Self-Monitoring~\cite{StevenWGangestad2000SelfmonitoringAA} & - & - & - & - & - & - & - & - & 18.04 & 5.67 & 48 & 35 \\
         Reinforced Cross-Modal~\cite{XinWang2018ReinforcedCM} & 10.65 & 3.53 & 67 & - & 11.46 & 6.09 & 43 & - & 11.97 & 6.12 & 43 & 38 \\
         EnvDrop~\cite{HaoTan2019LearningTN} & 11.00 & 3.99 & 62 & 59 & 10.70 & 5.22 & 62 & 48 & 11.66 & 5.23 & 51 & 47 \\
         AuxRN~\cite{FengdaZhu2020VisionLanguageNW} & - & 3.33 & 70 & 67 & - & 5.28 & 55 & 50 & - & 5.15 & 55 & 51 \\
         PREVALENT~\cite{hao2020towards} & \textbf{10.32} & 3.67 & 69 & 65 & 10.19 & 4.71 & 58 & 53 & 10.51 & 5.30 & 54 & 51 \\
         PRESS~\cite{XiujunLi2019RobustNW} & 10.35 & 3.09  & 71 & 67 & 10.06 & 4.31 & 59 & 55 & 10.52 & 4.53 & 57 & 53 \\
         AirBERT~\cite{PierreLouisGuhur2021AirbertIP} &11.09 &2.68 &75 &70 &11.78 &4.01 &62 &56 &12.41 &4.13 &62 &57 \\
         \hline
         VLN $\circlearrowright$ BERT\cite{hong2021vln} & 11.13 & 2.90 & 72 & 68 & 12.01 & 3.93 & 63 & 57 & 12.35 & 4.09 & 63 & 57 \\
         \textbf{GeoVLN (Ours)}  & 11.98 & 3.17 & 70 & 65 & 11.93 & 3.51 & 67 & 61 & 13.02 & 4.04 & 63 & 58 \\
         \hline
         HAMT~\cite{HAMT} &11.15 & 2.51 & 76 & 72 &11.46 &\textbf{2.29} &66 & 61 & 12.27 &\textbf{3.93} &\textbf{65} & 60 \\ 
         \textbf{GeoVLN$^{\dagger}$ (Ours)}  & 10.68 & \textbf{2.22} & \textbf{79} & \textbf{76} & 11.29 & 3.35 & \textbf{68} & \textbf{63} & 12.16 & 3.95 & \textbf{65} & \textbf{61} \\
         \bottomrule
    \end{tabular}
    }
    \caption{Comparison of \textbf{OUR MODEL} with the previous state-of-the-art methods on R2R dataset. $^{\dagger}$ indicates the results with HAMT as the backbone. The primary metric is SPL.}
    \label{tab:leaderboard}
\end{table*}

\section{Experiments}

\subsection{Experimental setup}
\label{sec:setup}

\noindent
{\bf Dataset}\quad
We use R2R~\cite{R2R} dataset for training and evaluation. The R2R dataset is built on 90 real-world indoor environments where the agents should traverse multiple rooms in a building to reach the destinations. And the navigation tasks are specifically described by 7189 trajectories and the corresponding instructions with the average length of 29 words. The dataset is divided into four sets including train, val seen, val unseen and test unseen sets, which mainly focus on the generalization capability of navigation in unseen environments.

\vspace{0.8em}
\noindent
{\bf Evaluation Metrics}\quad
We adopt the standard metrics used in previous works for evaluation:
1) Trajectory Length (TL): the average navigational trajectory length in meters; 
2) Navigation Error (NE): the distance between the final position of the agent and the target;
3) Success Rate (SR): the ratio of agents eventually stopping within 3 meters of the destination; and
4) Success Weighted by Path Length (SPL)~\cite{PeterAnderson2018OnEO} 
: SR weighted by the inverse of TL which measures how closely a trajectory aligns with the shortest path. 
A higher SPL score indicates a better balance between achieving the goal and taking the shortest path.

\vspace{0.8em}
\noindent
{\bf Implementation Details}\quad
Our multimodal visual features (including RGB, depth and surface normal features) are extracted by the pretrained CLIP-Res50x4 model~\cite{AlecRadford2021LearningTV}. In the mixture of imitation learning and reinforcement learning training process, $\lambda$ is set to be $0.2$. For fair comparisons, we follow the training pattern of the Recurrent VLN-BERT by mixing the original training data and the augmented data with 1:1 ratio. The experiments are performed on a single GeForce GTX TITAN X GPU with AdamW optimizer. To stabilize the gradient and accelerate the convergence, we use cosine annealing scheduler with warmup and set the maximum learning rate to $10^{-5}$. 
We train the network for 100,000 iterations with the batch size of 8 and then choose the model with the highest SPL on the validation unseen split for testing.

\begin{table*} \small
    \centering
    \renewcommand\arraystretch{1.0}
    \scalebox{1.0}{
    \begin{tabular}{cccccccc}
         \toprule
         \multicolumn{1}{c}{} & \multicolumn{3}{c}{Input} & \multicolumn{2}{c}{Val Seen} & \multicolumn{2}{c}{Val Unseen} \\
         \cmidrule(lr){2-4} \cmidrule(lr){5-6} \cmidrule(lr){7-8} 
         Model & RGB & DEPTH & NORMAL  & SR$\uparrow$ & SPL$\uparrow$ & SR$\uparrow$ & SPL$\uparrow$ \\
         \hline
         Baseline & \checkmark & & & \textbf{69.83} & 64.21 & 64.50 & 58.35 \\
         Baseline & \checkmark & \checkmark & & 66.99 & 63.28 & 63.86 & 58.58 \\
         Baseline & \checkmark & & \checkmark & 68.46 & 63.66 & 62.71 &  57.20\\
         Baseline & \checkmark & \checkmark & \checkmark & 66.41 & 62.51 & 64.75 & 59.31 \\
         \hline
         LSA & \checkmark &   &   & 67.58 & 63.06 & 64.62 & 59.78 \\
         LSA & \checkmark & \checkmark & \checkmark & 68.66 & 63.92 & 66.54 & 60.62 \\
         LSA + MAtt  & \checkmark &   &   & 68.46  & 63.92  & 66.11  & 60.31  \\
         LSA + MAtt (Full)  & \checkmark & \checkmark & \checkmark & 69.64 & \textbf{64.86} & \textbf{66.75} & \textbf{61.00} \\
         \bottomrule
    \end{tabular}
    }
    \caption{Ablation study on multi-modal visual inputs and LSA module with VLN $\circlearrowright$ BERT as the backbone.}
    \label{tab:ablation}
\end{table*}

\begin{figure*}
\vspace{-0.1in}
\begin{centering}
\includegraphics[width=0.82\linewidth]{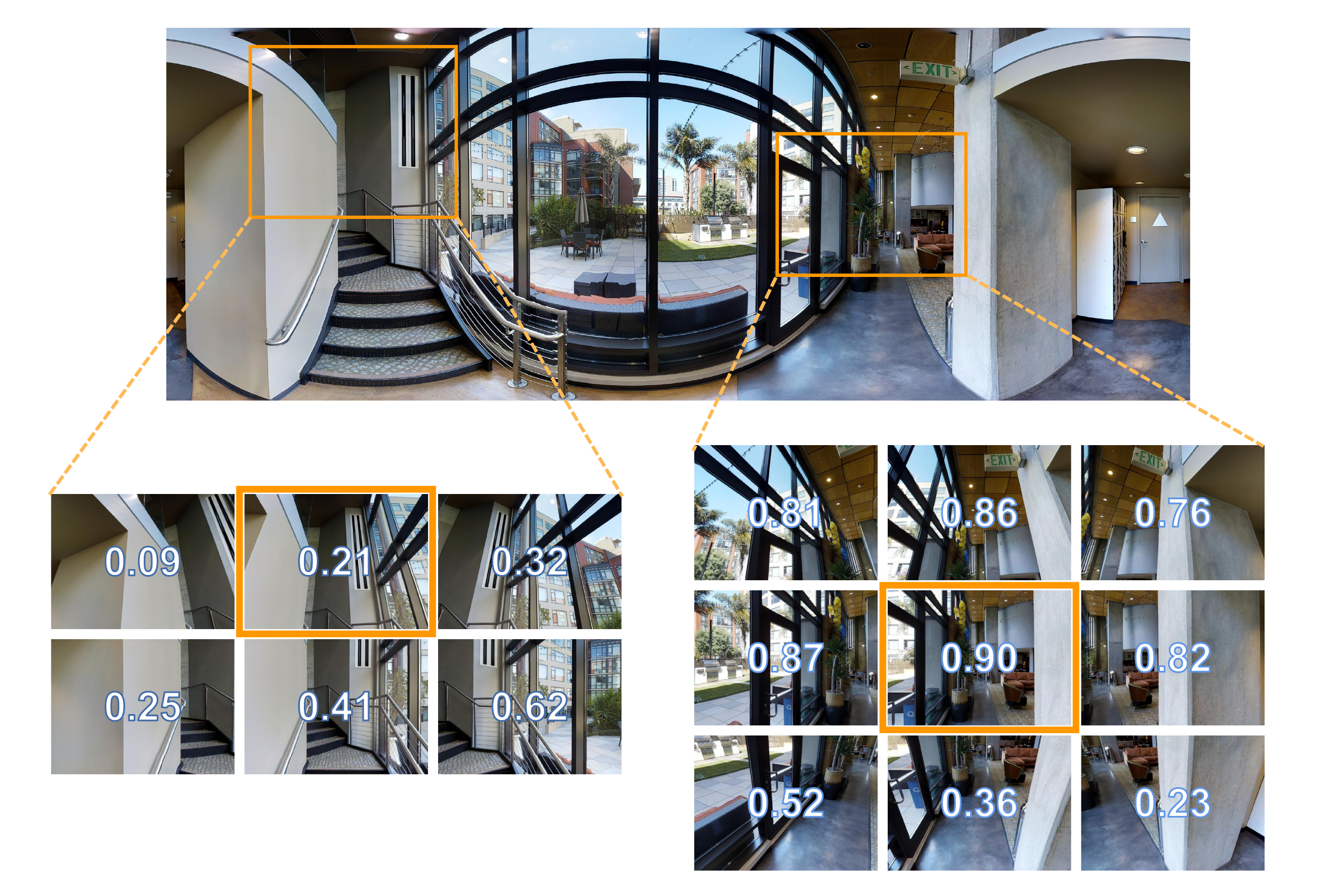}
\par\end{centering}
\vspace{-0.1in}
 \caption{An visualization example of Walk up stairs. It shows  the effectiveness of our local-aware slot attention module. \label{fig:ex1}}
\vspace{-0.1in}
\end{figure*}

\subsection{Main Results}
The main goal of R2R VLN task is to make optimal choice at each viewpoint based on past and current information, and find the best path towards target.
In this section, we provide the comparisons with previous works to show the effectiveness of the proposed GeoVLN.

\noindent \textbf{Competitors}. As baselines, we choose all the methods reported in \cite{hong2021vln} with the additional recent work AirBERT \cite{PierreLouisGuhur2021AirbertIP} and HAMT \cite{HAMT}. 
In addition, we extend HAMT with our newly designed modules (\ie GeoVLN$^{\dagger}$) to test the effectiveness of these modules, as our contributions in GeoVLN is actually orthogonal to \cite{HAMT}. 
Specifically, we directly use the Two Stage Visual Representation Learning Module to augment the original RGB features in the HAMT, and replace the original fully connected layer with our Multiway Attention Module to make decision. Further details on how we incorporate these modules into the HAMT model can be found in the Supplementary Material.

The quantitative results are shown in \cref{tab:leaderboard}. We mainly focus on the scores of SR and SPL in unseen environments, which provide a comprehensive evaluation of the generalization capability. Additionally, we also report the metrics of TL and NE.

Our results, presented in \cref{tab:leaderboard}, demonstrate the effectiveness of our proposed models based on VLN $\circlearrowright$ BERT and HAMT as the backbone network. Notably, our models achieve the best performance overall, outperforming all baseline methods, with particularly impressive results in unseen environments. In comparison to VLN $\circlearrowright$ BERT \cite{hong2021vln} on which our framework is built, GeoVLN improves SPL and SR by 7.0\% and 6.3\%, respectively, on the val-unseen split. Similarly, when compared to HAMT, our proposed modules lead to significant improvements of 3.3\% and 3.0\% on SPL and SR, respectively. These results demonstrate the efficacy of our GeoVLN approach. While our performance is slightly inferior to VLN $\circlearrowright$ BERT on the val-seen split, our experimental results support our claims and contributions.

Further, our geometry-enhanced visual representation is derived from object-centric learning \cite{slotatt}. Unfortunately, we notice that most of previous works can only successfully work the slot attention mechanism on synthetic dataset, \eg CLEVR3D. In contrast, we extend it to VLN task to encourage feature fusion between spatially neighboring views. It is capable of working under the complex real-world environments of R2R dataset. 

Furthermore, to  reveal the insights,  we provide an ablation study with visualization results about our local slot attention in the following subsections.

\subsection{Ablation Study}
In this section, we provide extensive ablation studies to validate the effectiveness of novel technical designed in our GeoVLN. 
For fair comparisons, all the variants in our experiment follow the same training setup described in \cref{sec:setup}.

Our quantitative results with VLN $\circlearrowright$ BERT as the backbone are presented in \cref{tab:ablation}. The ``baseline'' denotes VLN $\circlearrowright$ BERT trained with the RGB features extracted by CLIP as the visual representation.

Firstly, we change the composition of different types of visual inputs, the results show that merely add depth map or normal map cannot offer better performance. This is possibly because the estimated depth and normal may contain errors which do not perfectly match RGB captures, so that hinder test accuracy. 
However, when both depth and normal inputs are given, they can benefit each other and provide geometry information that facilitates navigation. This is evidenced by improved performance in the Baseline, LSA, and LSA+MAtt models when depth and normal inputs are provided, highlighting the effectiveness of multi-modal visual inputs like depth and normal.

Next, we demonstrate the efficacy of our local-aware slot attention (LSA) and multiway attention (MAtt) by adding them to the baseline model one by one. 
The results show that the inclusion of LSA improves SPL by 2.5\% on the val-unseen split.
And by incorporating our MAtt module, our full model facilitates the identification of the most relevant visual modality for different phrases, resulting in superior performance compared to the baseline. Notably, MAtt provides valuable interpretability for decision-making processes, as demonstrated in the Supplementary material.

\subsection{Visualization}
To further show the effectiveness of our local-aware slot attention module, we show an visualization example in \cref{fig:ex1}. 
The panoramic image above shows the whole room, and we choose two candidate view (in orange bounding box) for visualization below. The number on each image denotes attention score.

As shown on the left, the agent arrives at the location of stairs, and it needs information about stairs and handrail as reference to make decision of next move. So the nearby images containing stairs or handrail have higher attention score, which means that the agent successfully obtains useful features from local neighbors to aid decision-making. 
More visualization results are shown in Supplementary material, which illustrate the effectiveness of both our local-aware slot attention module and multiway attention module.

\section{Conclusions}
This paper introduces GeoVLN, which learns \textbf{Geo}metry-enhanced visual representation based on slot attention for robust \textbf{V}isual-and-\textbf{L}anguage \textbf{N}avigation. We compensate RGB captures with the estimated depth maps and normal maps as visual observations, and design a novel two-stage slot-based module
to learn geometry-enhanced visual representation. Moreover, a multiway attention module is presented to facilitate decision-making. Extensive experiments on R2R dataset demonstrate the effectiveness of our newly designed modules and show the compelling performance of the proposed method.

\noindent\textbf{Acknowledgement}. {This work was supported by STMP of Commission of Science and Technology of Shanghai (No.21XD1402500), and Shanghai Municipal STMP (2021SHZDZX0103). }

{\small
\bibliographystyle{ieee_fullname}
\bibliography{egbib}
}

\clearpage
\noindent \textbf{\Large Supplementary Material}
\vspace{0.1in}
\setcounter{section}{0}


\section{Details of the expansion on HAMT}
Here, we describe how to extend HAMT with our Two Stage Visual Representation Learning Module and Multiway Attention Module.

At each time step, HAMT adopts the instruction $\boldsymbol{I}$, current observations $\boldsymbol{O}_t$, and historical observations $\boldsymbol{H}_t$ to handle environment information. We augment the original RGB observations using the two-stage visual representation learning module. This involves using local-aware slot attention and multimodal fusion to act upon the current candidate observations, while the other observations, both historical and current, are only augmented with multimodal fusion. As a result, we obtain the geometrically enhanced visual representations $\hat{\boldsymbol{F}}_t$ and $\hat{\boldsymbol{H}_t}$ for current observations $\boldsymbol{O}_t$ and historical observations $\boldsymbol{H}_t$, respectively. The backbone network of HAMT, which includes the unimodal and cross-modal transformer encoders, is then used to obtain the embeddings $\boldsymbol{I}^{\prime}, \hat{\boldsymbol{H}_t^{\prime}}, \hat{\boldsymbol{F}_t^{\prime}}$, which can be formulated as:
\begin{equation}
\begin{split}
\boldsymbol{I}^{\prime}, \hat{\boldsymbol{H}_t^{\prime}}, \hat{\boldsymbol{F}_t^{\prime}}&=\operatorname{HAMT}\left(\boldsymbol{I}, \hat{\boldsymbol{H}_t}, \hat{\boldsymbol{F}}_t\right).
\end{split}
\end{equation}
Since HAMT does not maintain a sufficiently information-rich state vector 
$\boldsymbol{s}_t$ as RecBERT does, we slightly adapt the MAtt module to fit HAMT's original network framework. Specifically, we use the $\text{[CLS]}$ token of the instruction to obtain a state vector, but first multiply the $\text{[CLS]}$ token and the features $\hat{\boldsymbol{F}_t^{\prime}}$ of each observation before calculating a matching score for each modality in the same way as the main text. 
The reformulated $\boldsymbol{s}_t$ can be written as: 
\begin{equation}
\begin{split}
\boldsymbol{s}_t=\boldsymbol{I}_{cls}^{\prime} \odot \hat{\boldsymbol{F}_t^{\prime}},
\end{split}
\end{equation}
where $\odot$ is element-wise multiplication,  $\boldsymbol{I}_{cls}^{\prime}$ is the embedding of $\text{[CLS]}$ token of the instruction. This process is consistent with the decision-making module in HAMT and also compensates for the visual information that is difficult to include in the $\boldsymbol{I}_{cls}^{\prime}$.

\section{More Training Details}

We train the network with RecBERT as the backbone for 100,000 iterations on a single GeForce GTX TITAN X GPU, and the one with HAMT as the backbone for 200,000 iterations on a single GeForce RTX 3090 GPU. The batchsize is set to 8. The optimizer is AdamW and a cosine annealing scheduler with warmup is used to adjust the learning rate. We set the arguments of the cosine annealing scheduler as follows: first cycle step size is $50$, cycle steps magnification is $1$, max learning rate is $10^{-5}$, min learning rate is $5 \times 10^{-8}$, decrease rate of max learning rate by cycle is $0.1$. The scheduler is employed to adjust the learning rate at an interval of 2000 iterations. 

\section{Computation Cost}

We compare the parameters, inference time, and memory usage of our extended models with those of the original models. The inference time is measured as the single-run time on the val unseen split. Although our extended models introduce additional computation cost, the cost is manageable compared to the improvement achieved. Importantly, our GeoVLN model achieves competitive performance with the original HAMT model on the val unseen split while incurring lower computational cost.
\begin{table}[h] \small
    \centering
    \renewcommand\arraystretch{1.0}
    \scalebox{0.9}{
    \begin{tabular}{cccccccc}
         \toprule
         \multicolumn{1}{c}{Model} & \multicolumn{1}{c}{Params (M)} & \multicolumn{1}{c}{Memory (GiB)} & \multicolumn{1}{c}{Inference Time (s)} \\
         \hline
         RecBERT & 153 & 6.0 & 70  \\
         GeoVLN & 159 & 7.8 & 108  \\
         HAMT & 163 & 9.9 & 134  \\
         GeoVLN$^{\dagger}$ & 174 & 15.1 & 191  \\
         \bottomrule
    \end{tabular}
    }
    \caption{Comparison of the computation cost. GeoVLN uses RecBERT as the backbone, while GeoVLN$^{\dagger}$ uses HAMT as the backbone.}
    \label{tab:comparison}
\end{table}

\section{Structure Variants of Two-Stage Module}

\begin{figure*}
  \centering
  \begin{subfigure}{0.24\linewidth}
    \includegraphics[width=0.99\linewidth]{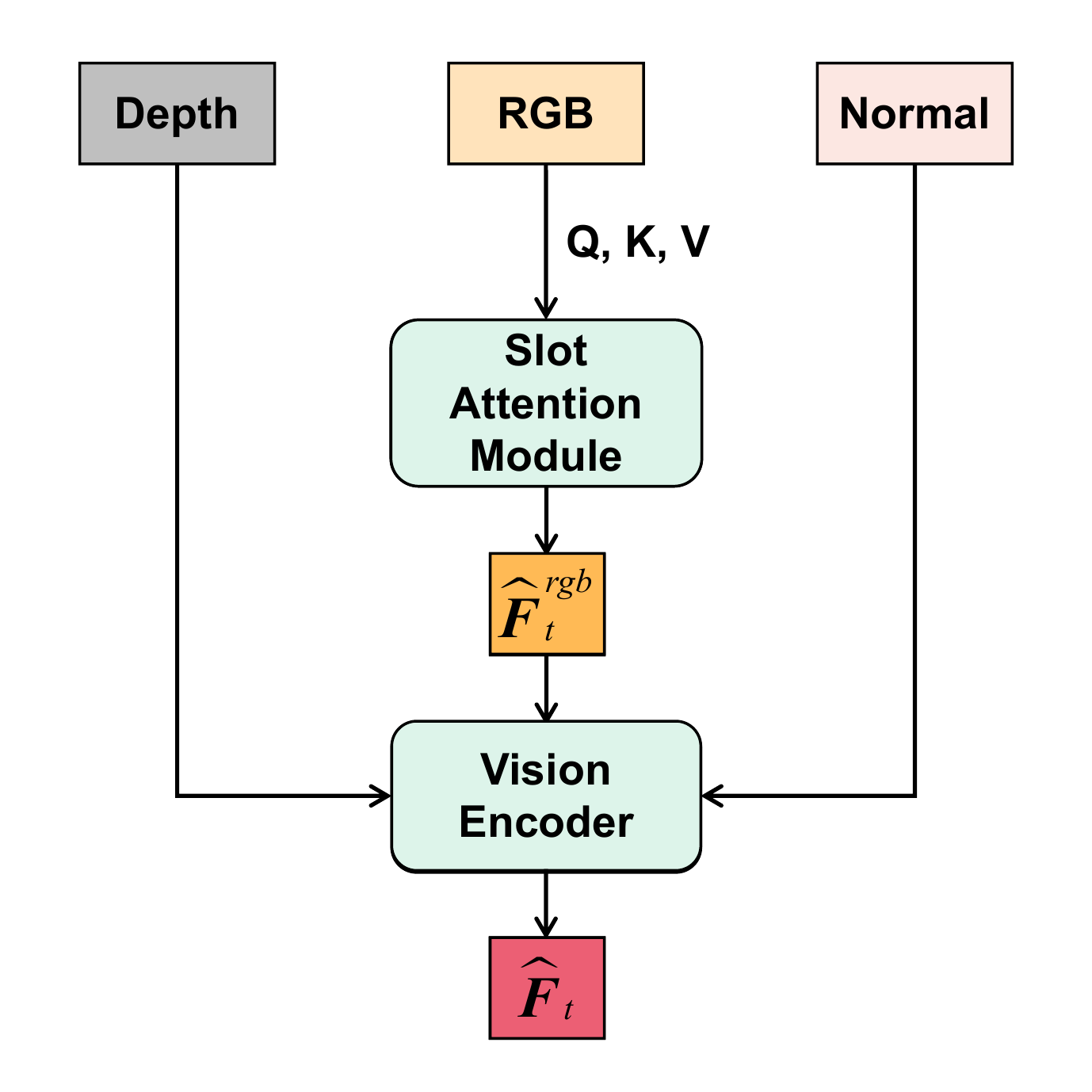}
    \caption{TwoSM}
    \label{fig:twostage-a}
  \end{subfigure}
  \hfill
  \begin{subfigure}{0.24\linewidth}
    \includegraphics[width=0.99\linewidth]{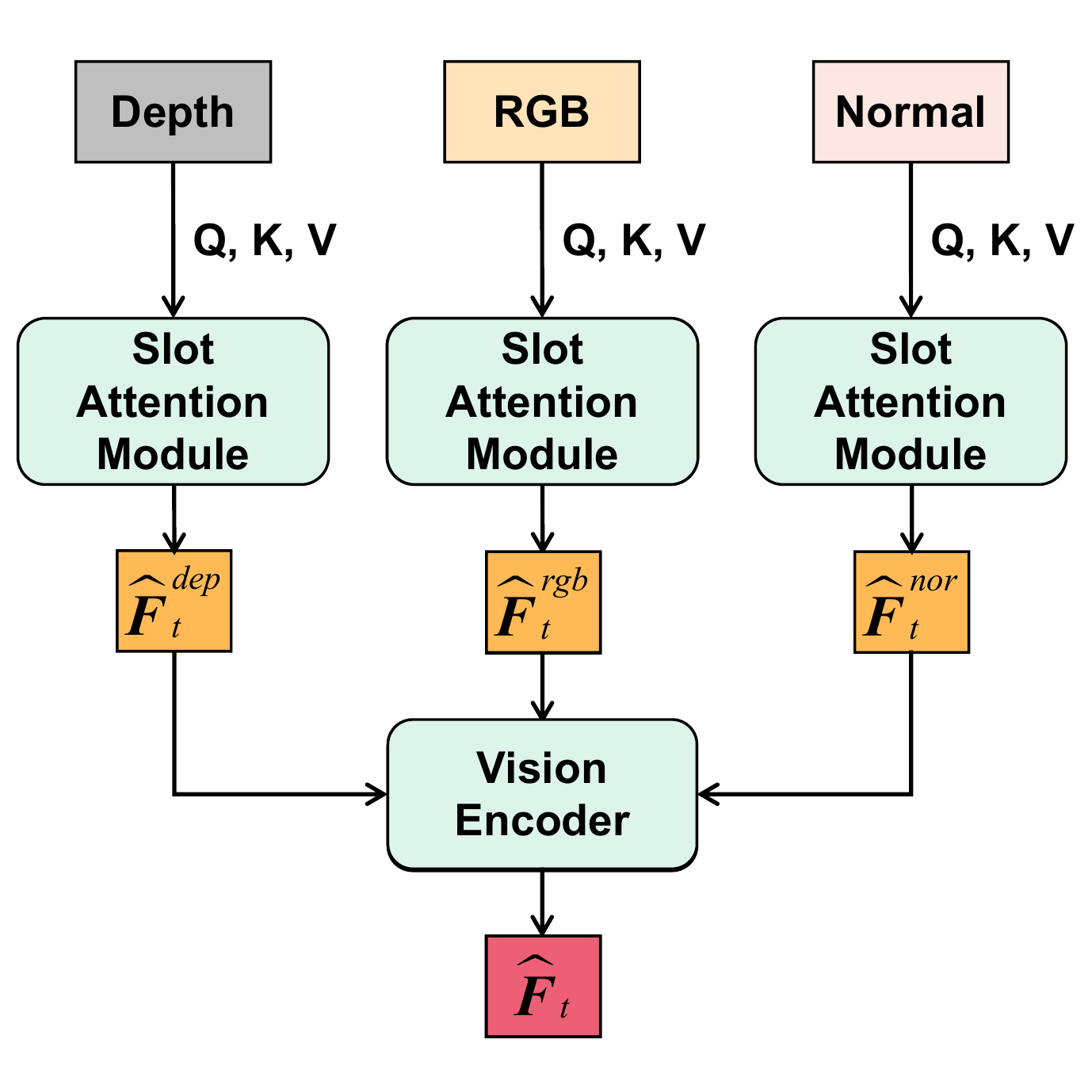}
    \caption{TwoSM-1}
    \label{fig:twostage-b}
  \end{subfigure}
  \hfill
  \begin{subfigure}{0.24\linewidth}
    \includegraphics[width=0.99\linewidth]{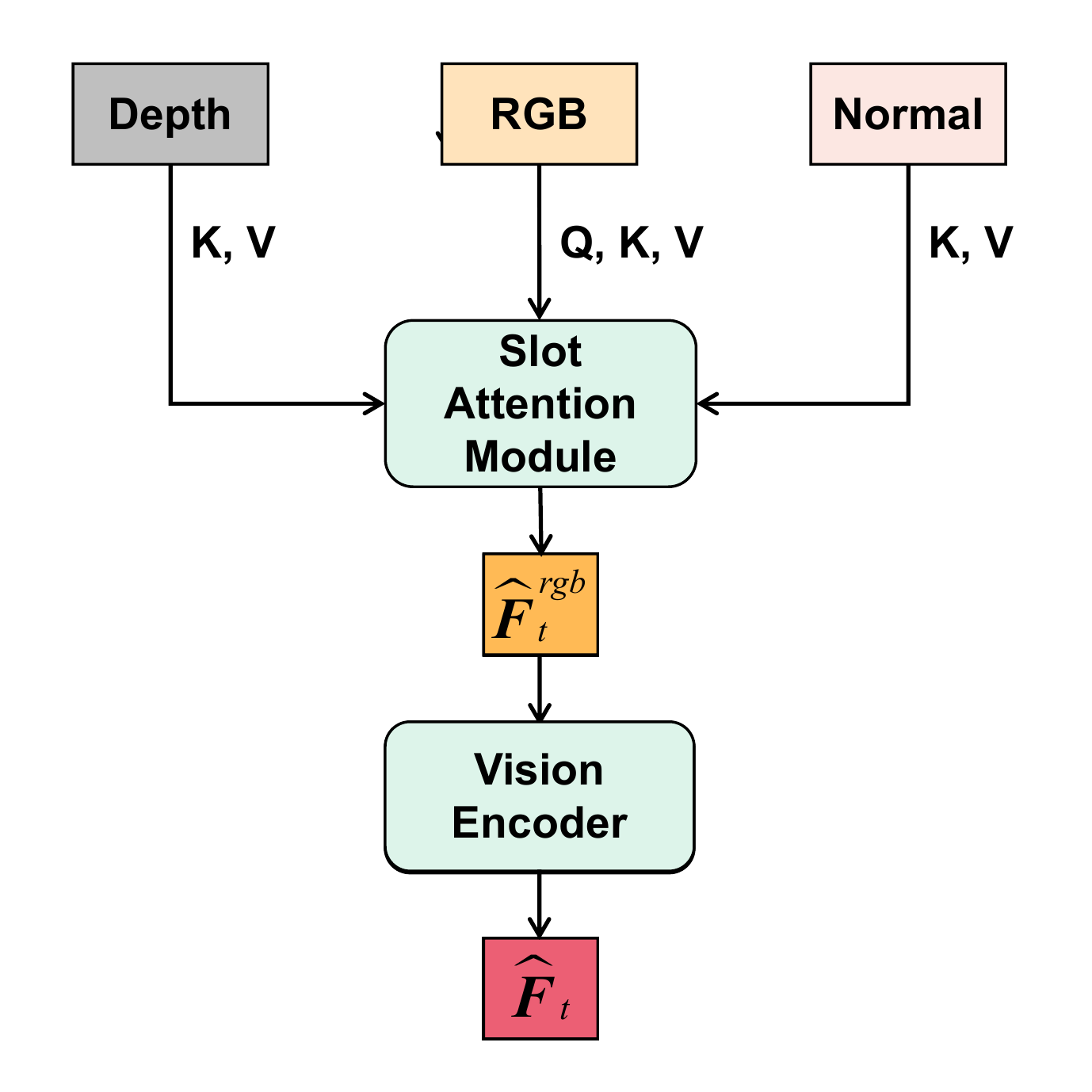}
    \caption{TwoSM-2}
    \label{fig:twostage-c}
  \end{subfigure}
  \hfill
  \begin{subfigure}{0.24\linewidth}
    \includegraphics[width=0.99\linewidth]{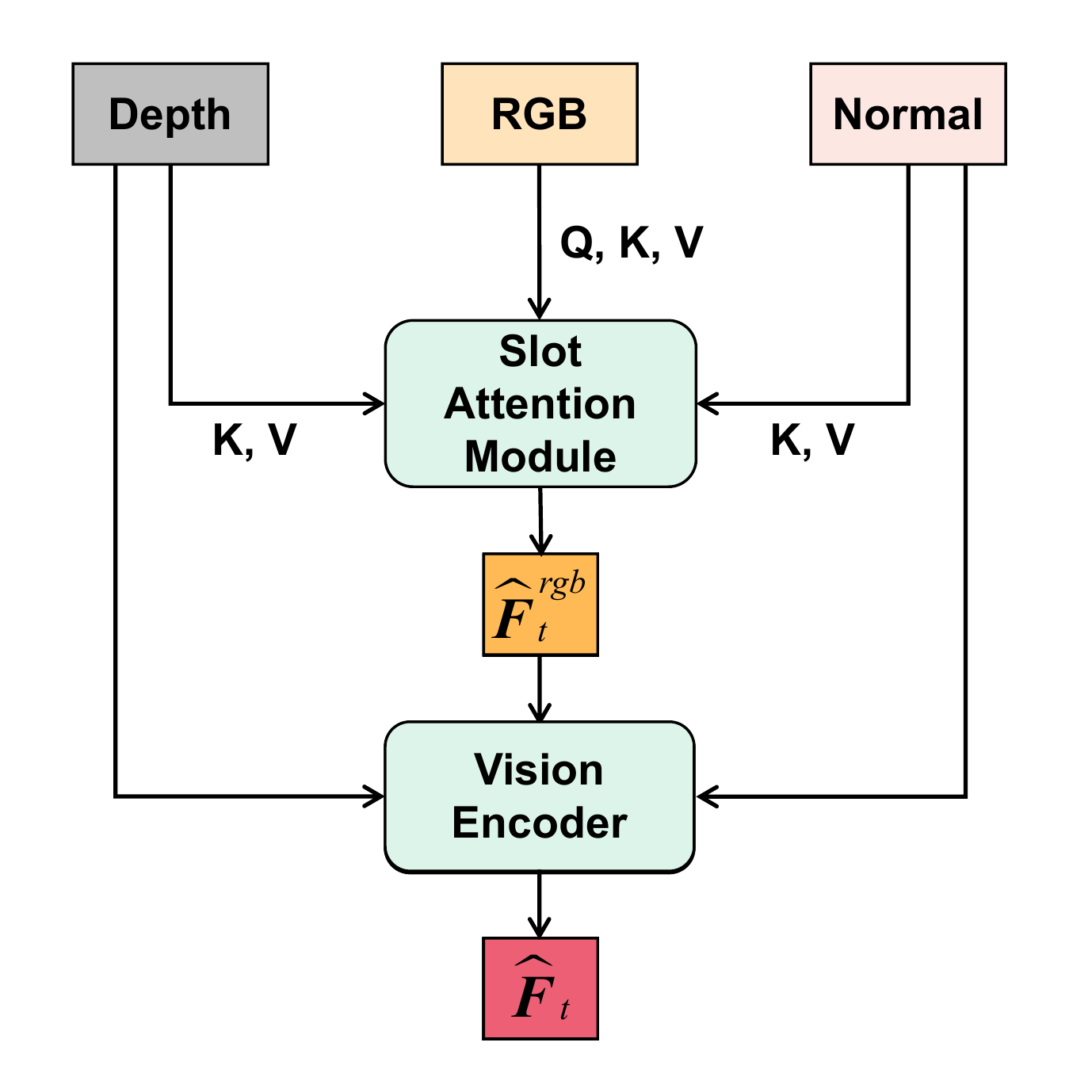}
    \caption{TwoSM-3}
    \label{fig:twostage-d}
  \end{subfigure}
  \caption{Pipelines of the Two-Stage Model and its structure variants.}
  \label{fig:twostage}
\end{figure*}

In this paper, we introduce a two-stage module to handle the multimodal observations including RGB, depth and normal surface. To further explore the better architecture of the two-stage model, we design three variants (\cref{fig:twostage-b,fig:twostage-c,fig:twostage-d}) of it and compare them with the original one (\cref{fig:twostage-a}). 

The two-stage model (TwoSM) combines a local-aware slot attention module and a vision encoder to acquire a geometry-enhanced visual representation $\hat{\boldsymbol{F}}_t$. The pipeline of the original one, which is adopted in our article, is shown in \cref{fig:twostage-a}. We firstly aggregate RGB features with local-aware slot attention and obtain the slot enhanced RGB features $\hat{\boldsymbol{F}}_t^{rgb}$. And then, the enhanced features $\hat{\boldsymbol{F}}_t^{rgb}$ together with the depth features and normal surface features are fed into the vision encoder, in which they are concatenated and projected to the geometry-enhanced visual representations $\hat{\boldsymbol{F}}_t$. Based on this method, three variants, named TwoSM-1, TwoSM-2 and TwoSM-3, are proposed. As shown in \cref{fig:twostage-b}, TwoSM-1 aggregate the RGB, depth and surface normal information respectively by a slot attention module, and then feeds the features $\hat{\boldsymbol{F}}_t^{rgb}$, $\hat{\boldsymbol{F}}_t^{dep}$, $\hat{\boldsymbol{F}}_t^{nor}$ to the vision encoder, performing the same operations as the original one. Moreover, we design a RGB-guided slot attention module for the second and third variants. In this module, slots are initialized with only RGB candidate features to dominate the fusion process, while multimodal features of the nearby observations are treated as both keys and values. Such a method enables the depth and surface normal information from nearby observations to directly update the slots (\ie the RGB candidate features) as well. TwoSM-2 (\cref{fig:twostage-c}) and TwoSM-3 (\cref{fig:twostage-d}) differ only in their visual encoders. TwoSM-2 takes only the enhanced RGB information $\hat{\boldsymbol{F}}_t^{rgb}$ as input, while TwoSM-3 additionally adding the original depth and normal information as input.

We quantitatively compare the three variants with the original TwoSM in \cref{tab:ablation}. It can be seen that all variants perform weaker than our original TwoSM. The best performer among the variants is TwoSM-1, but with a 1.5\% reduction in SPL and 3.0\% reduction in SR on the val-unseen split. We suggest that this could indicate that the local information aggregation based on slot attention in depth and normal features instead interferes with the decision making of the agents. TwoSM-2 and TwoSM-3 additionally add depth and surface normal information to update RGB candidate features in slot module, but present worse performance. 
This suggests that the simultaneous aggregation of nearby observations and depth/normal maps presents a significant challenge. In particular, there exist certain features that do not correspond spatially or modally with the RGB-initialized slots, making it extremely challenging to align them with the slots. In addition, TwoSM-3 outperforms TwoSM-2, suggesting that adding raw depth and normal information to the visual encoder contributes to the model. The above results illustrate the rationality and superiority of our two-stage model. 

\begin{table} \small
    \centering
    \renewcommand\arraystretch{1.2}
    \scalebox{1.0}
    {
    \begin{tabular}{cccccccc}
         \toprule
         \multicolumn{1}{c}{} & \multicolumn{2}{c}{Val Seen} & \multicolumn{2}{c}{Val Unseen} \\
         \cmidrule(lr){2-3} \cmidrule(lr){4-5} 
         Model   & SR$\uparrow$ & SPL$\uparrow$ & SR$\uparrow$ & SPL$\uparrow$ \\
         \hline
         TwoSM  &  \textbf{69.64} & \textbf{64.86} & \textbf{66.75} & \textbf{61.00} \\
         TwoSM-1 &  65.72 & 62.23 & 64.75 & 60.06 \\
         TwoSM-2 &  67.29 & 63.27 & 62.15 & 56.63 \\
         TwoSM-3 & 67.19 & 63.14 & 64.75 & 59.46 \\
         \bottomrule
    \end{tabular}
    }
    \caption{Ablation study on the structure variants of the two-stage model.}
    \label{tab:ablation}
\end{table}

\section{Supplementary Results}

In this section, more results of the navigation path are given visually. Moreover, additional qualitative or quantitative results are given to illustrate the effectiveness of our local-aware slot attention module and multiway attention module.

\subsection{Visualization for Local-aware Slot Attention}

\begin{figure*}
  \centering
  \begin{subfigure}{1\linewidth}
    \includegraphics[width=0.99\linewidth]{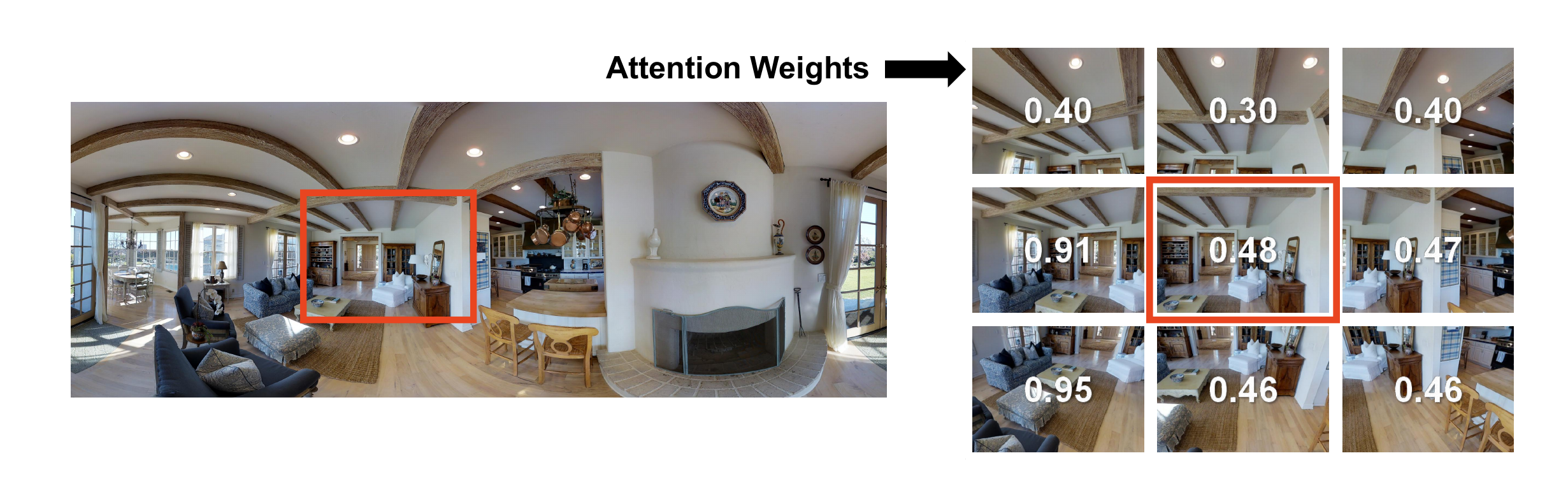}
    \caption{The instruction is ``Turn to your right and go pass the couch. Turn left and then turn right and stop by the couch''. According to the instruction ``couch'', the two regions containing the sofa on the left side are given significantly more attention than the other regions.}
    \label{fig:slot-ex1}
  \end{subfigure}
  \vfill
  \begin{subfigure}{1\linewidth}
    \includegraphics[width=0.99\linewidth]{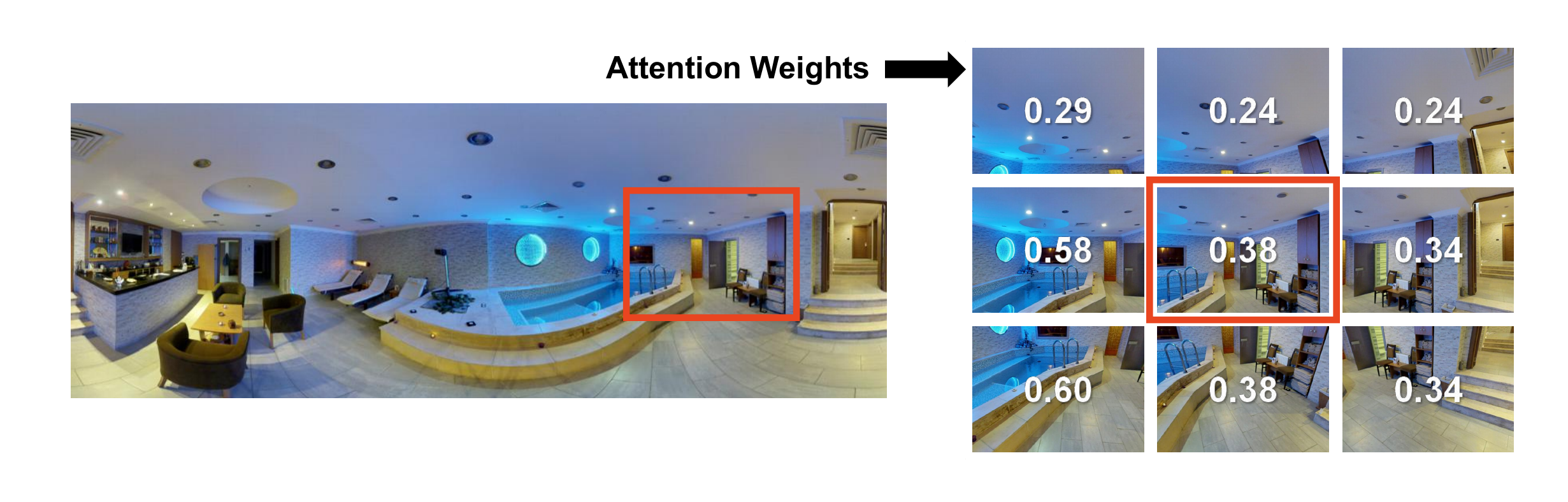}
    \caption{The instruction is ``Pass the pool then go into the beaded curtain room and turn left then wait there right at the entrance of the sauna''. The regions containing the pool are given greater attention weight, allowing the agent to correctly execute the instruction ``pass the pool".}
    \label{fig:slot-ex2}
  \end{subfigure}
  \vfill
  \begin{subfigure}{1\linewidth}
    \includegraphics[width=0.99\linewidth]{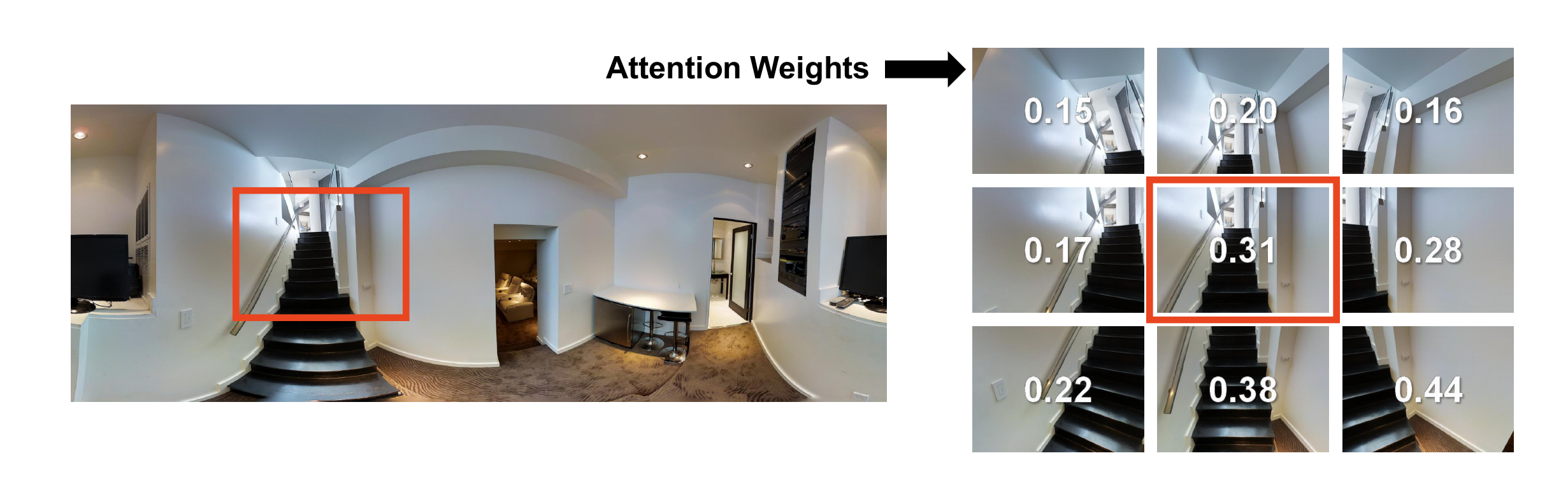}
    \caption{The instruction is ``Go up the stairs then turn right and stand near the bed''. The views containing stairs (bottom two rows) are given higher attention weights than the views with almost no stairs (top row).}
    \label{fig:slot-ex3}
  \end{subfigure}
  \caption{Visualizations of the attention weights in local-aware slot attention module.}
  \label{fig:slot-ex}
\end{figure*}

We show how the local-aware slot attention module aggregates local observations to candidate views in \cref{fig:slot-ex}. The left side of the figure shows a panoramic view, with the candidate view selected by our agent marked in red. The observations used to update that candidate view and the corresponding attention weights are labeled on the right side. Note that the updates obtained from the attention weights need to be added to the original candidate view. Thus, even if the candidate view does not have the highest attention score, the final candidate information still maintains the highest relevance to the candidate view itself. 

\begin{figure*}
\begin{centering}
\includegraphics[width=0.92\linewidth]{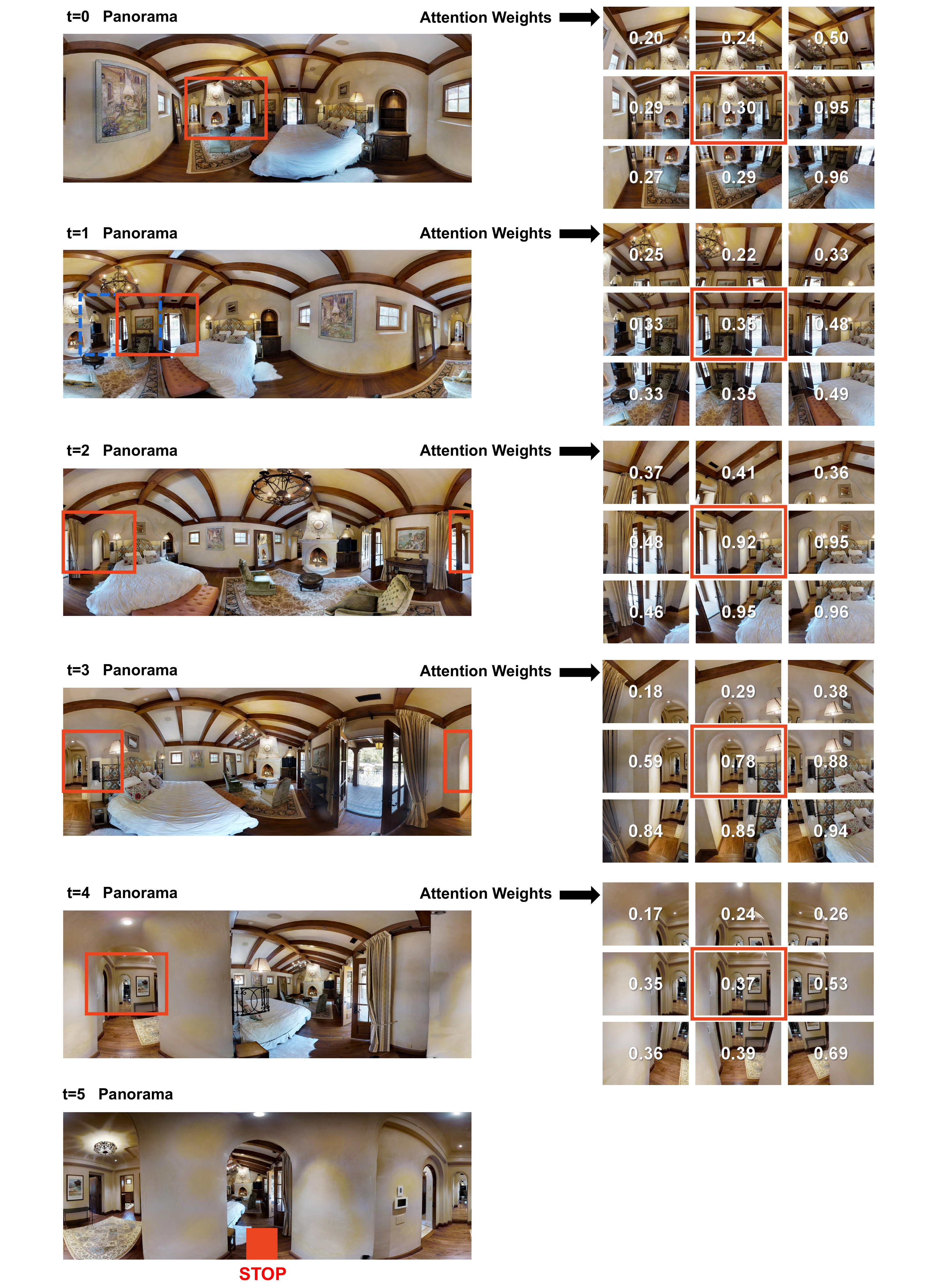}
\par\end{centering}
\vspace{-0.1in}
 \caption{An example of success on the val unseen split. The instruction is ``Go around the bed and to the right. Go through the arch opening and wait near the thermostat.''} \label{fig:traj3}
\vspace{-0.1in}
\end{figure*}

\cref{fig:traj3} shows an case of our GeoVLN successfully navigating while the Recurrent VLN-BERT does not in val-unseen set. The red box marks the candidate view we select (the successful one) and the blue box marks the candidate view that the Recurrent VLN-BERT selects incorrectly the first time. At the time steps $t=0 \sim 3$, our local-aware slot attention module gives high attention weights to the regions containing the bed (even greater than $0.9$). Therefore, our model finally succeeds in capturing the relationship between the bed and the candidate views and completes the instruction ``Go around the bed and to the right'', while the Recurrent VLN-BERT fails.

\begin{figure*}
\begin{centering}
\includegraphics[width=0.92\linewidth]{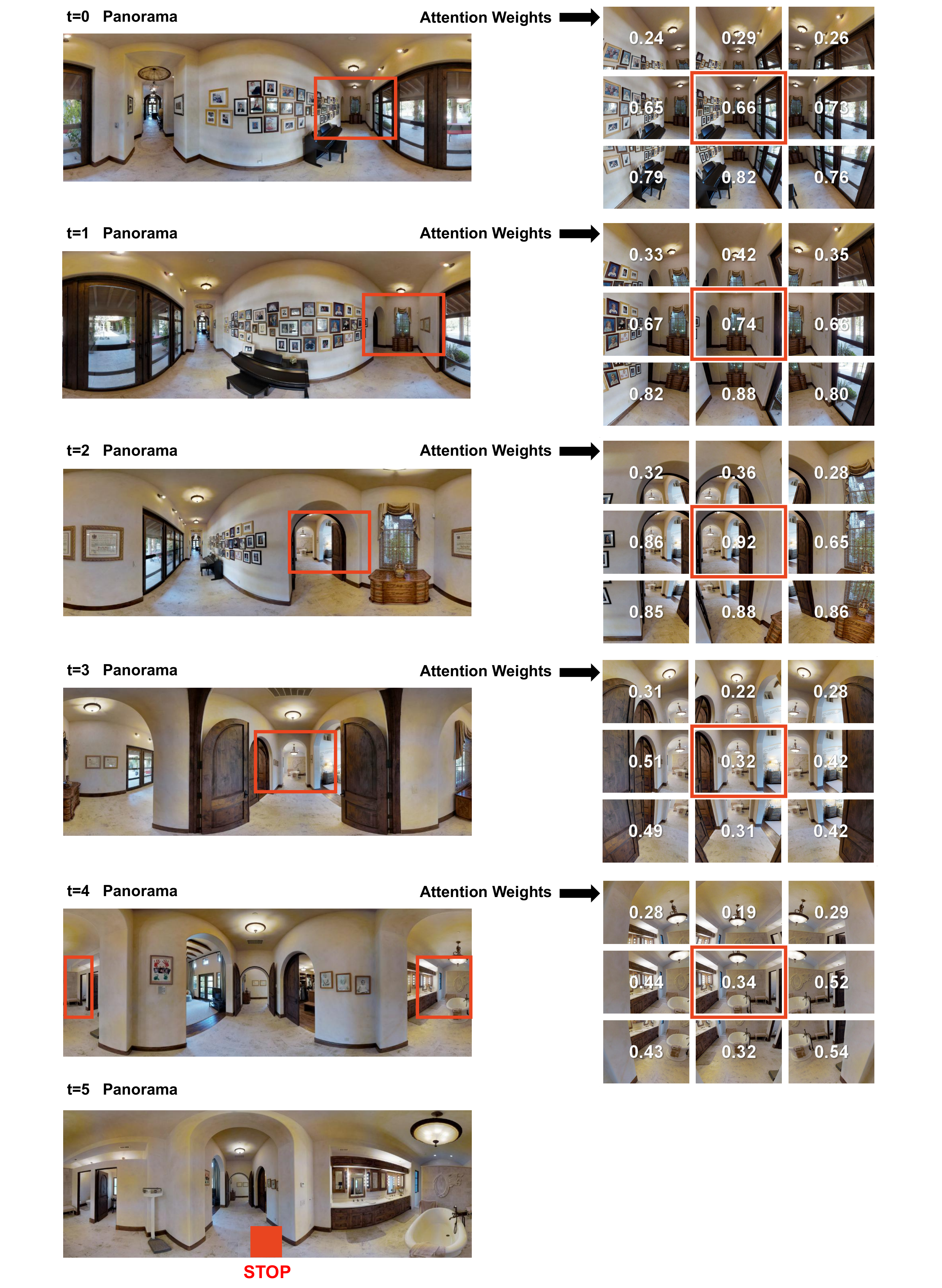}
\par\end{centering}
\vspace{-0.1in}
 \caption{An example of success on the val unseen split. The instruction is ``Walk past piano. Walk through arched wooden doors. Wait at bathtub.''} \label{fig:traj1}
\vspace{-0.1in}
\end{figure*}

\begin{figure*}
\begin{centering}
\includegraphics[width=0.92\linewidth]{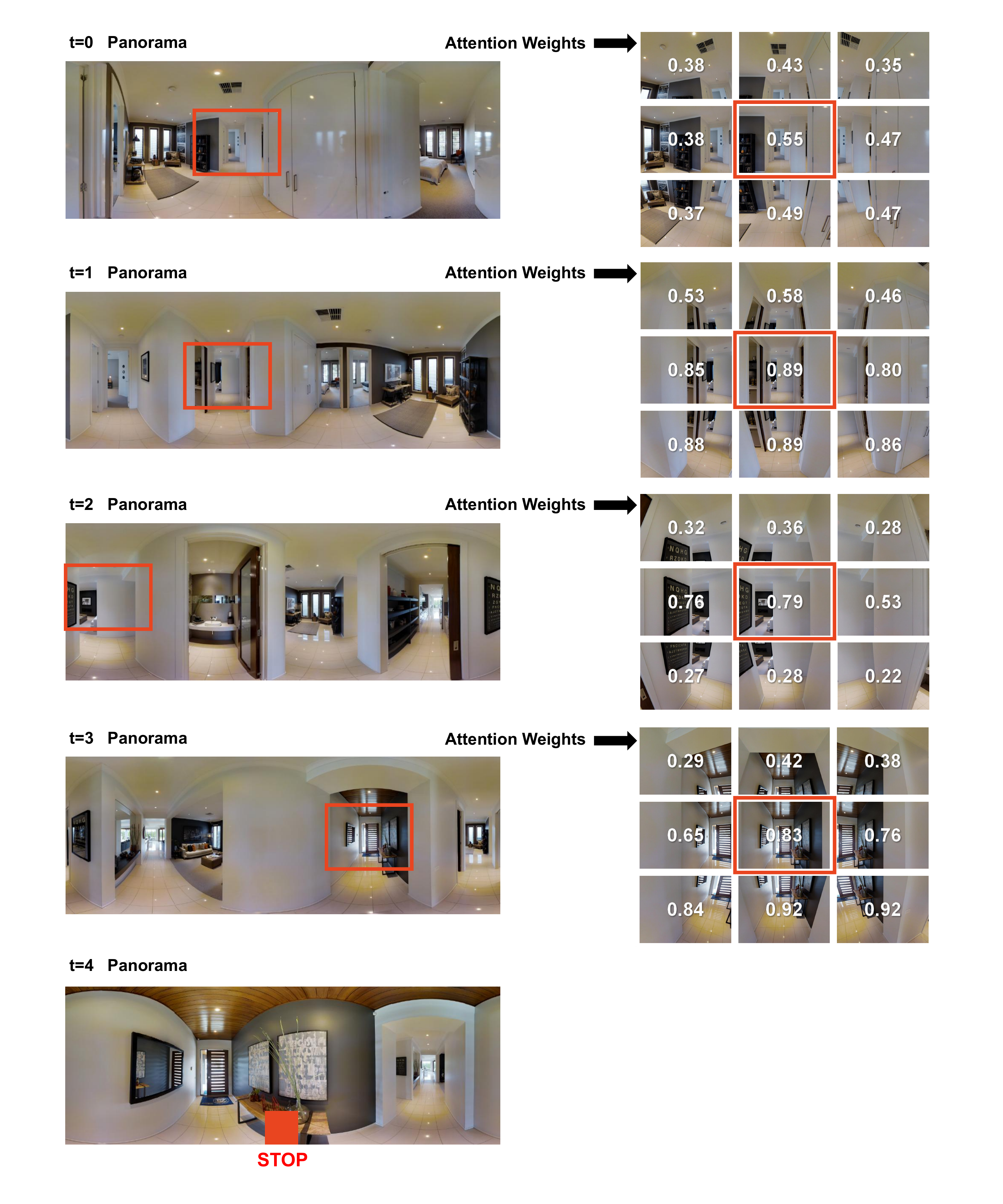}
\par\end{centering}
\vspace{-0.1in}
 \caption{An example of success on the val unseen split. The instruction is ``Leave the bedroom and take the first right into a hallway. Take a right at the end of the hall and enter the foyer. Stop before you reach the mirror.''}
 \label{fig:traj2}
\vspace{-0.1in}
\end{figure*}

\cref{fig:traj1,fig:traj2} show additional cases of the success of our model in the val-unseen spit, with the attention weights for local observations marked on the right. The piano and the wooden door are given more attention in \cref{fig:traj1} and the hall is given more attention in \cref{fig:traj2}, which are consistent with the instructions.

\subsection{Visualization for Multiway Attention}

\begin{figure*}
\begin{centering}
\includegraphics[width=0.9\linewidth]{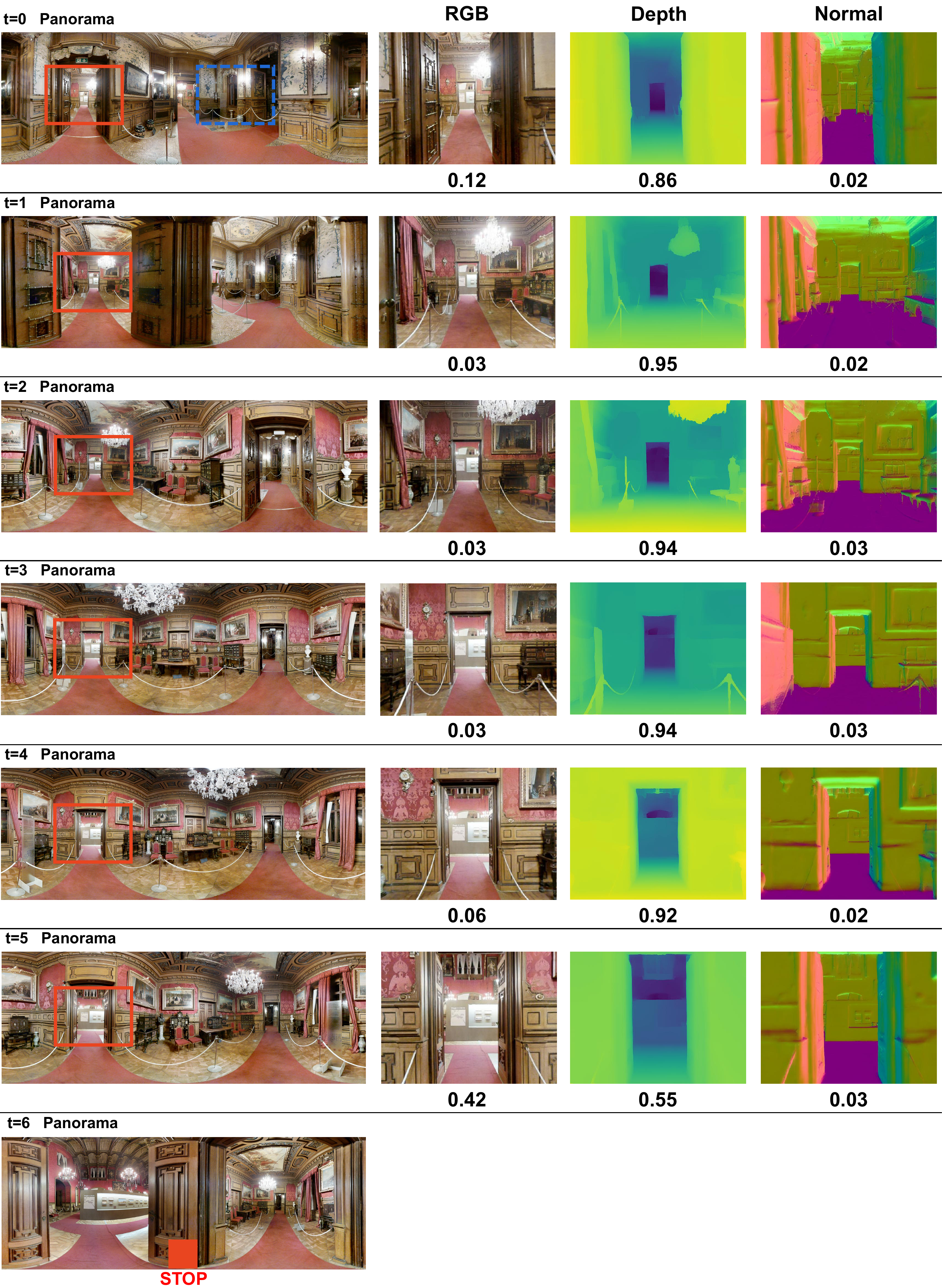}
\par\end{centering}
\vspace{-0.1in}
 \caption{An example of success on the val unseen split. The instruction is ``Follow the red carpet through the double doors. Continue straight through the room and wait in the doorway with the double doors at the end.'' The red box marks the choice of GeoVLN and the blue box marks the wrong choice of the Recurrent VLN-Bert. The numbers below the RGB, depth, and surface normal images indicate the weight coefficients, including $w_t^{rgb}$, $w_t^{dep}$ and $w_t^{nor}$, given by the multiway attention module, respectively.}
 \label{fig:matt}
\vspace{-0.1in}
\end{figure*}

In the multiway attention module, the final matching score is obtained by dynamically weighting the matching scores of the three modalities of visual inputs (RGB, depth and normal images) and the instructions. \cref{fig:matt} shows how the weight coefficients of the three modalities ($w_t^{rgb}$, $w_t^{dep}$, $w_t^{nor}$) change in a single successful navigation.
In this case, the depth information rather than the RGB information is given the highest weight coefficient as the instruction contains ``straight through the room''. At initialization ($t=0$), our model correctly selects the candidate view containing the ``double doors'' (marked by the red box), while the Recurrent VLN-BERT incorrectly selects the view containing the window (marked by the blue box). This is attributed to the fact that the depth map contains depth information about the room behind the door, making the agent aware of the door. In addition, at the time step $t=5$, the weight coefficient $w_t^{rgb}$ increases to $0.42$ and $w_t^{dep}$ decreases to $0.55$. This is consistent with the instruction to ``wait in the doorway with the double doors at the end", which requires higher attention to the RGB images to confirm the appearance of the door.

\subsection{Failure Cases}

\begin{figure*}
  \centering
  \begin{subfigure}{0.48\linewidth}
    \includegraphics[width=0.99\linewidth]{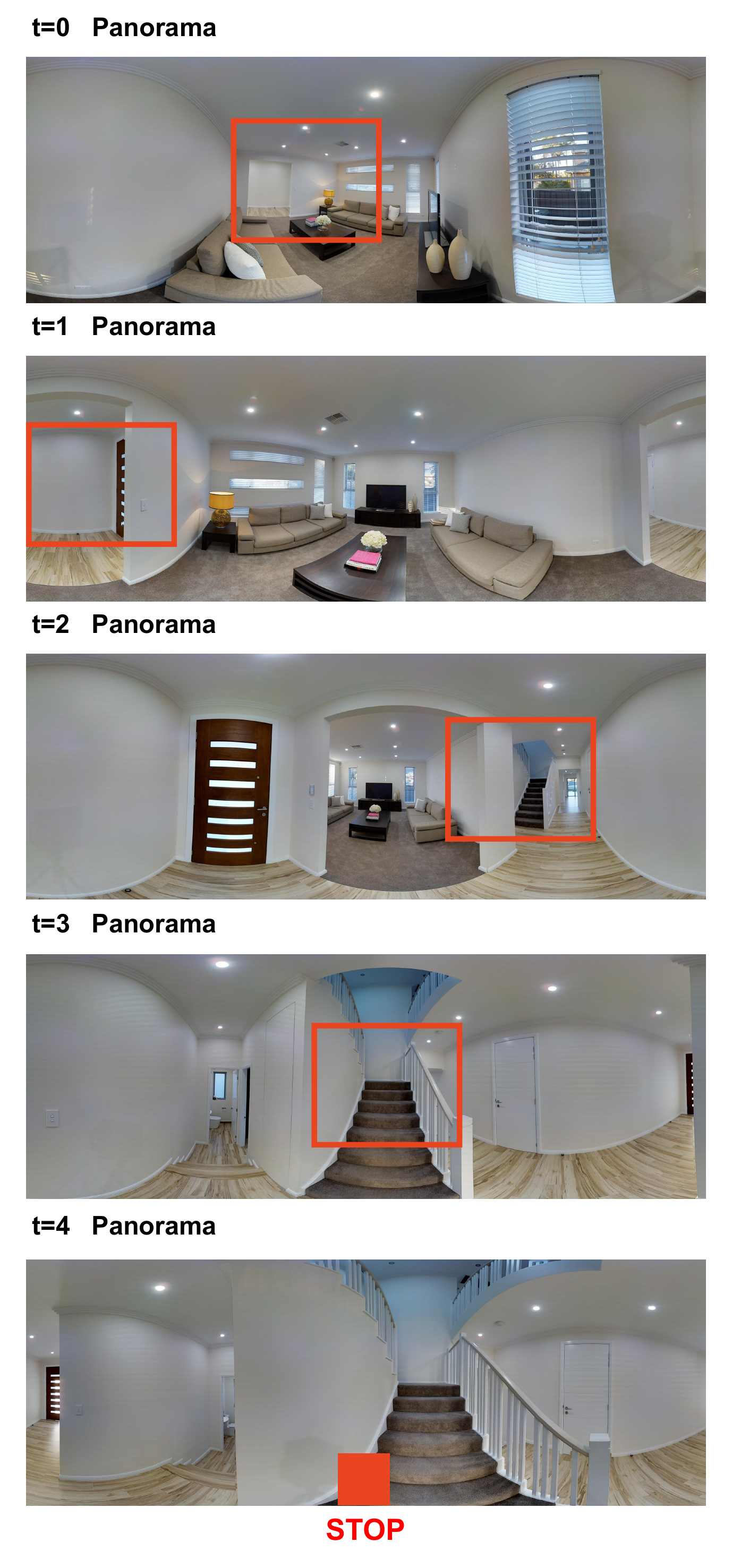}
    \caption{Ground truth trajectory.}
    \label{fig:fail1-gt}
  \end{subfigure}
  \hfill
  \begin{subfigure}{0.48\linewidth}
    \includegraphics[width=0.99\linewidth]{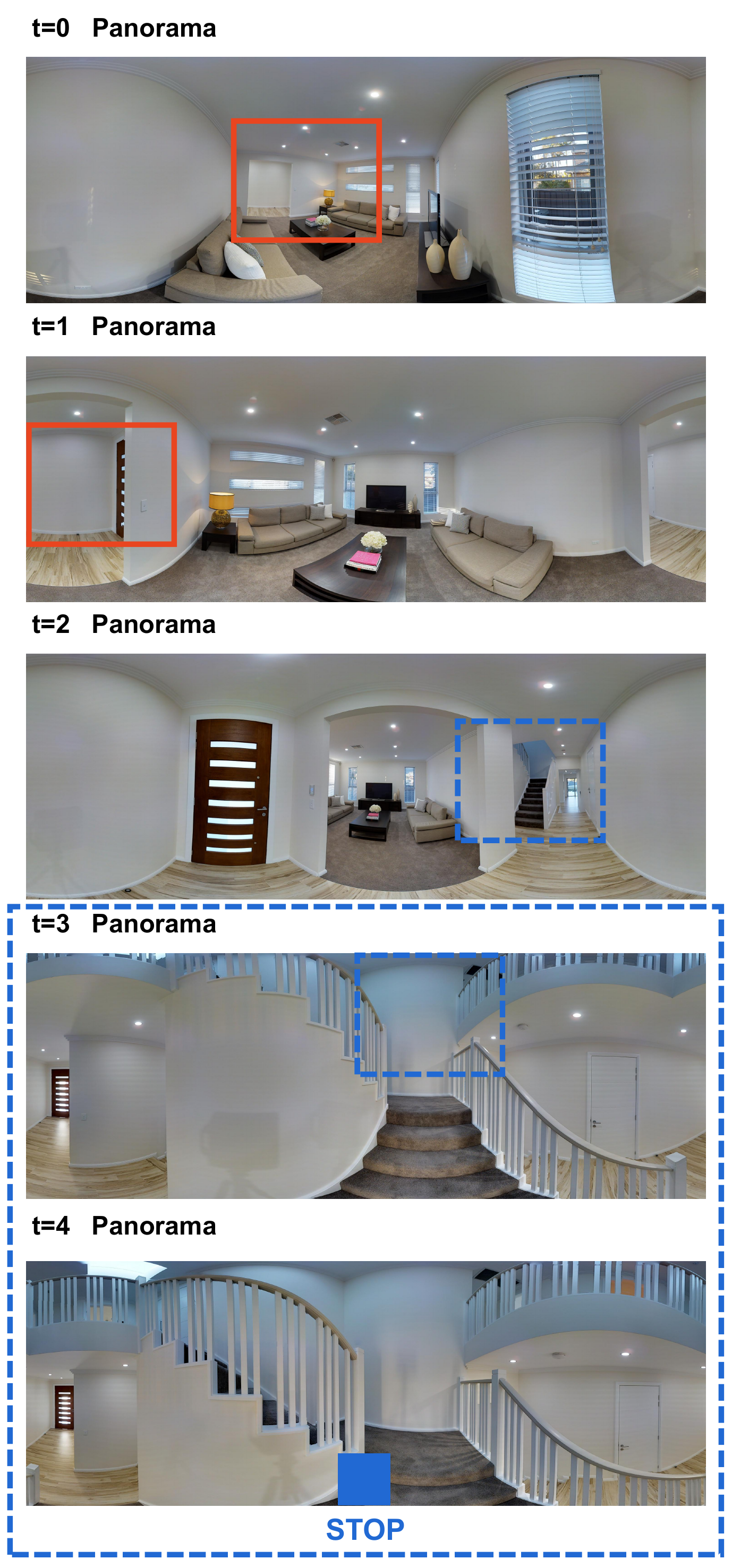}
    \caption{Predicted trajectory.}
    \label{fig:fail1-pre}
  \end{subfigure}
  \caption{An example of failure to navigate on the val unseen split. The instruction is ``Walk through the bedroom and out into the hall way. Turn left and walk up to the stairs. Walk up to the first step and stop''. The blue boxes indicate incorrect selections. Our GeoVLN fails to correctly perform the instruction "Walk up to the first step and stop" and instead continues up the stairs.}
  \label{fig:fail1}
\end{figure*}

\begin{figure*}
  \centering
  \begin{subfigure}{0.48\linewidth}
    \includegraphics[width=0.99\linewidth]{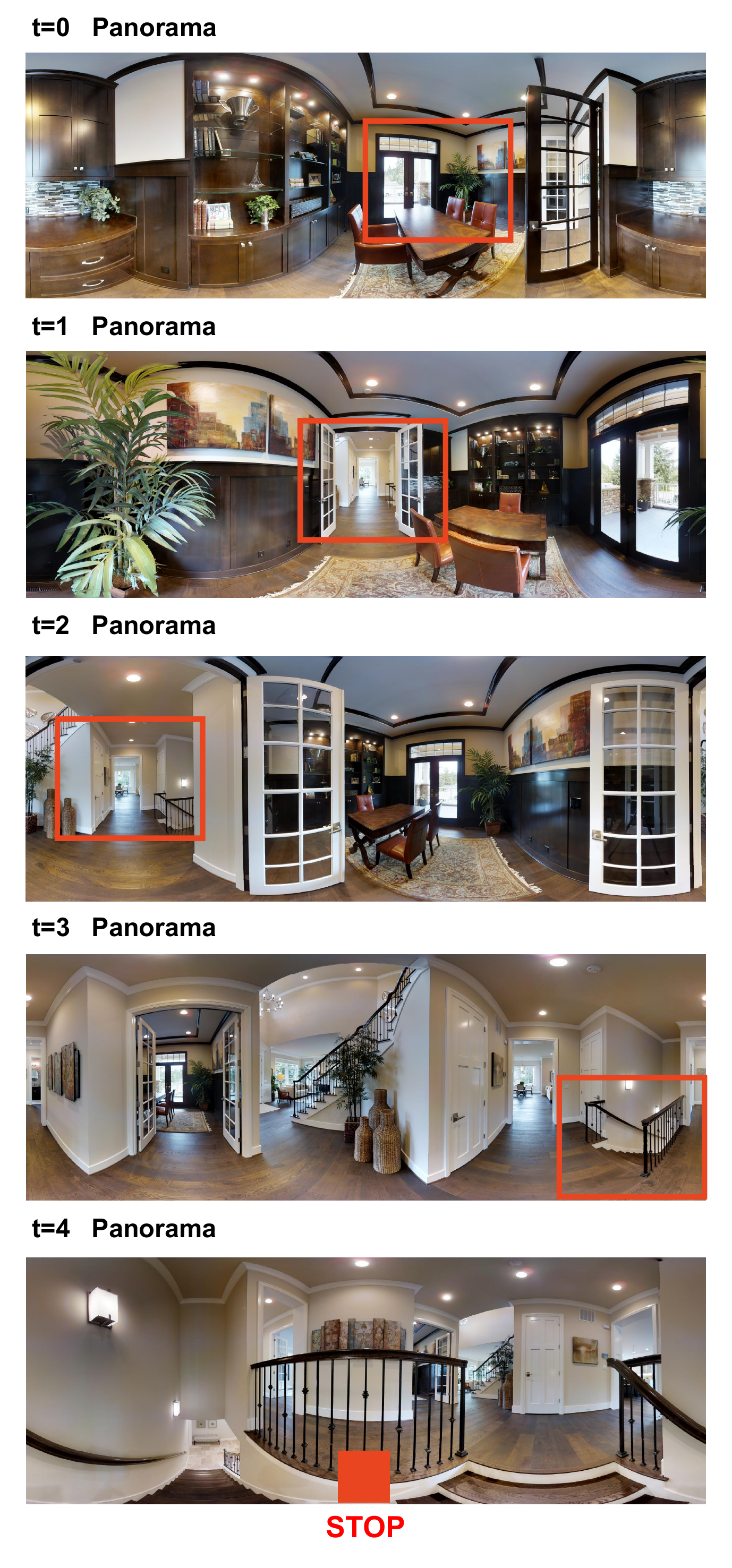}
    \caption{Ground truth trajectory.}
    \label{fig:fail2-gt}
  \end{subfigure}
  \hfill
  \begin{subfigure}{0.48\linewidth}
    \includegraphics[width=0.99\linewidth]{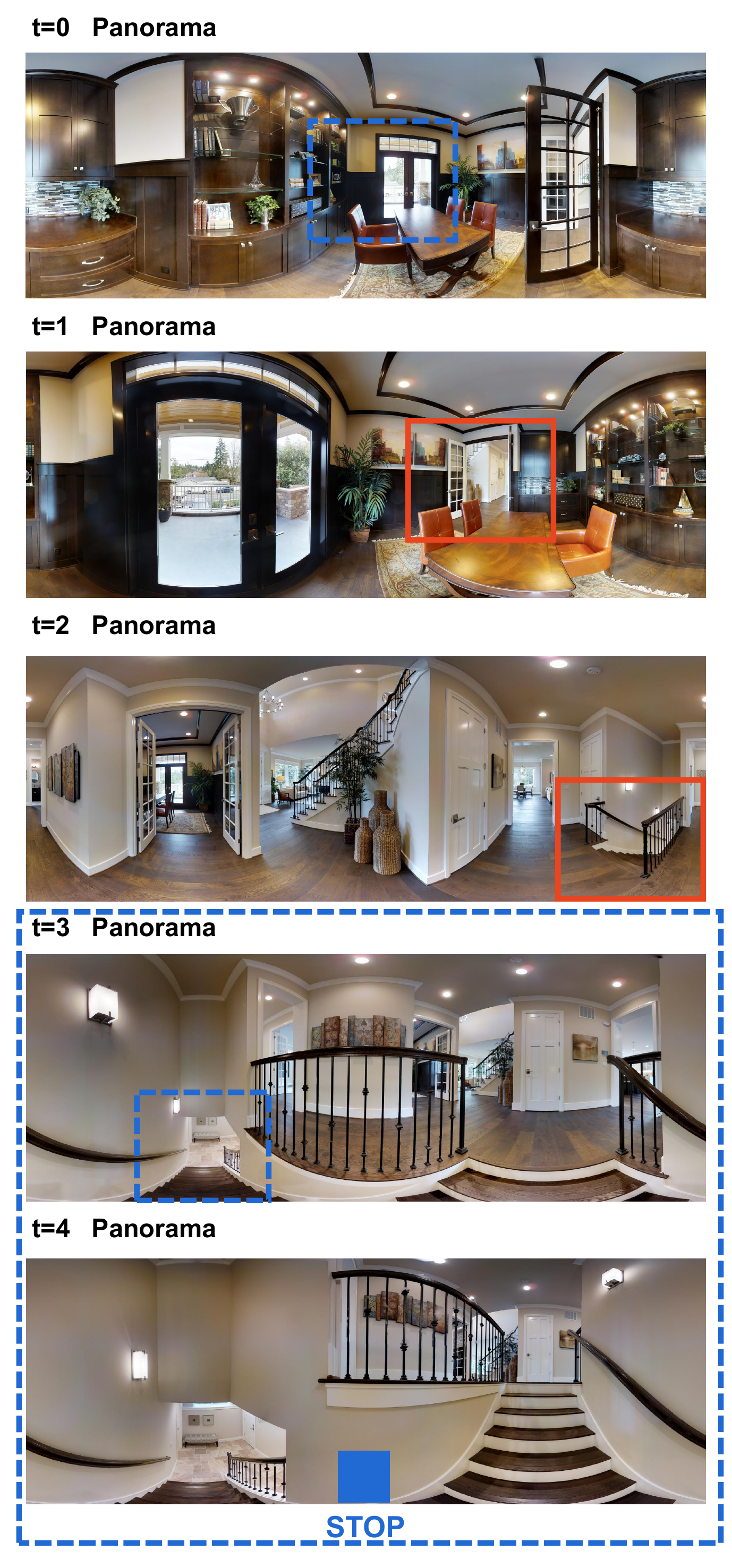}
    \caption{Predicted trajectory.}
    \label{fig:fail2-pre}
  \end{subfigure}
  \caption{An example of failure to navigate on the val unseen split. The instruction is ``Leave the room by exiting through the open double doors. Go down the stairs and stop on the second step from the top and wait there''. The blue boxes indicate incorrect selections. At initialization ($t=0$), the direction chosen by GeoVLN is deviated, but subsequently this mistake is rectified ($t=1, 2$). However, at the end, the agent does not stop correctly on the stairs, but continues downward instead.}
  \label{fig:fail2}
\end{figure*}

Examples of navigation failures of GeoVLN are shown in \cref{fig:fail1,fig:fail2}, which demonstrate that our model still has limitations. More accurate and effective models for vision-and-language navigation are expected to be proposed.

\end{document}